\pdfoutput=1
\documentclass[10pt,twocolumn,amsmath,amssymb,longbibliography,nofootinbib,showkeys,eprint]{revtex4-2}

\usepackage[activate={true,nocompatibility},final,tracking=true,kerning=true,spacing=true]{microtype}
\usepackage{epigraph}
\usepackage{graphicx}
\usepackage{epstopdf}
\usepackage{braket}
\usepackage{bm}
\usepackage{numprint}
\usepackage{float}
\usepackage[T2A,T1]{fontenc}
\usepackage[utf8]{inputenc}
\usepackage{comment}
\usepackage{hhline}
\usepackage{xfrac}
\usepackage{amsthm}
\usepackage{mathtools}
\usepackage{makecell}
\usepackage{dsfont}
\usepackage{calc}
\usepackage{seqsplit}
\usepackage{circuitikz}
\usepackage{hyperref}
\hypersetup{
    colorlinks=true
}
\usepackage{orcidlink}

\begin{document}
\preprint{V.M.}
\title{Semidefinite Programming for Quantum Channel Learning}

\author{Mikhail Gennadievich \surname{Belov}}
\email{mikhail.belov@tafs.pro}
\affiliation{Lomonosov Moscow State University,  Faculty of Mechanics and Mathematics,
   GSP-1,  Moscow, Vorob'evy Gory, 119991, Russia}
\affiliation{Autretech Group, Skolkovo Innovation Center, Nobel Street, Building 7, Moscow, 121205, Russia}

\author{Victor Victorovich \surname{Dubov}}
\email{dubov@spbstu.ru}
\affiliation{Peter the Great St. Petersburg Polytechnic University, 195251, Russia}

\author{Vadim Konstantinovich \surname{Ivanov}\,\orcidlink{0000-0002-3584-4583}}
\email{ivvadim@rambler.ru}
\affiliation{Peter the Great St. Petersburg Polytechnic University, 195251, Russia}

\author{Alexander Yurievich \surname{Maslov}\,\orcidlink{0009-0005-6296-5988}}
\email{maslov.ton@mail.ioffe.ru}
\affiliation{Ioffe Institute, Politekhnicheskaya 26, St Petersburg, 194021, Russia}

\author{Olga Vladimirovna \surname{Proshina}\,\orcidlink{0000-0003-0087-5838}}
\email{proshina.ton@mail.ioffe.ru}
\affiliation{Ioffe Institute, Politekhnicheskaya 26, St Petersburg, 194021, Russia}

\author{Vladislav Gennadievich \surname{Malyshkin}\,\orcidlink{0000-0003-0429-3456}} 
\email{malyshki@ton.ioffe.ru}
\affiliation{Ioffe Institute, Politekhnicheskaya 26, St Petersburg, 194021, Russia}

\date{December, 19, 2025}

\begin{abstract}
\begin{verbatim}
$Id: SemidefiniteProgrammingQuantumChannel.tex,v 1.174 2026/09/10 14:14:58 mal Exp $
\end{verbatim}
The problem of reconstructing a quantum channel from a sample of classical data is considered.
When the total fidelity can be represented as a ratio of two quadratic forms
(e.g., in the case of mapping a mixed state to a pure state, projective operators, unitary learning, and others),
Semidefinite Programming (SDP) can be applied to solve the fidelity optimization problem with respect to the Choi matrix.
A remarkable feature of SDP is that the optimization is convex,
which allows the problem to be efficiently solved by a variety of numerical algorithms.
We have tested several commercially available SDP solvers, all of which allowed for the reconstruction of quantum channels
of different forms. A notable feature is that the Kraus rank of the obtained quantum channel typically
comprises less than a few percent of its maximal possible value.
This suggests that a relatively small Kraus rank quantum channel is typically sufficient
to describe experimentally observed classical data.
The theory was also applied to the problem of reconstructing projective operators from data.
Finally, we discuss a classical computational model based on quantum channel transformation,
performed and calculated on a classical computer, possibly hardware-optimized.
\bigskip \\ \noindent DOI: \href{https://doi.org/10.1103/kdl5-smrh}{10.1103/kdl5-smrh} Phys. Rev. E \textbf{114}, 035302 (2026)
\end{abstract}
\maketitle


\section{\label{intro}Introduction}
Knowledge representation is a crucial and distinctive feature in any Machine Learning (ML) technique. The form of the knowledge determines the degree of generalization a technique can exhibit. More generally, the form of knowledge representation can be considered as a computational framework. For example, ML techniques such as statistical learning \cite{vapnik1974method},
logical approaches \cite{hajek1977generation}, support vector machines \cite{vapnik2013nature},
rules and decision trees \cite{witten2002data}, fuzzy logic \cite{zadeh1965fuzzy, hajek1995fuzzy},
and deep learning \cite{bengio2013representation} can all be viewed as forms of computation.
The learning process consists of building a model from the data, essentially solving a form of an inverse problem.

It is of special interest to relate computations to quantum mechanics.
The most widely discussed form is to correspond a computational model to an actual quantum system,
the time evolution of which would perform the computation (quantum computing).
Another approach is to correspond a computational model to a quantum system,
where knowledge representation is typically in the form of quantum channel mapping,
which is then computed on a classical computer.
This paper considers the latter approach.
The most general form of quantum evolution is a quantum channel, typically in the form of completely positive trace-preserving (CPTP) maps \cite{jamiolkowski1972linear, choi1975completely, kraus1983states, belavkin1986radon}.
\begin{align}
A^{OUT}&=\sum_{s=0}^{N_s-1} B_s A^{IN} B^{\dagger}_s
\label{KrausOperator}
\end{align}
The minimum number of terms in (\ref{KrausOperator}) is called the Kraus rank.
For $N_s=1$, it becomes a unitary quantum channel.
The use of unitary operators for ML/AI knowledge representation has attracted attention
recently\cite{bisio2010optimal,arjovsky2016unitary,hyland2017learning,huang2021learning,yu2023optimal,belov2024partiallyPRE,belov2024quantumPRE}.

The paper allows for a two-sided consideration: quantum channel learning in the quantum-information sense,
and a learning framework for classical data. Whereas in our previous works \cite{belov2024quantumPRE,belov2025superstatePRE}
we were mostly concerned with the algebraic problem corresponding to the quantum inverse problem,
here the inverse problem is reduced to an optimization problem that, under certain assumptions on the fidelity form, is convex.
This particularly explains why we were able to develop an iterative algorithm for finding
the global maximum solution to the unitary learning problem in \cite{belov2024partiallyPRE}.
The contributions on the quantum-information side mostly consist of demonstrating that the inverse problem
can be reduced to a convex optimization problem when the fidelity is expressed as a ratio of two quadratic forms
in mapping operators.
The contributions on the ML/AI side consist of generalizing the commonly studied
unitary learning framework\cite{bisio2010optimal,arjovsky2016unitary,hyland2017learning,huang2021learning,yu2023optimal,belov2024partiallyPRE,belov2024quantumPRE}
to the general form of a quantum channel (\ref{KrausOperator}),
as well as providing a practical demonstration of the applicability of the theory.
It is important that we can go beyond CPTP maps; for example, we can reconstruct projective operators.
This represents an important practical problem in ML/AI for knowledge representation, clustering, mapping identification,
and related tasks. Another result is the numerical demonstration that quantum channels reconstructed
from classical data typically have low Kraus rank,
which enables a number of optimizations.
Whether this low-Kraus-rank property also persists in actual quantum systems remains an open question.

In \cite{belov2025superstatePRE}, we introduced the concept of the density matrix network,
which considers a quantum channel as a classical computational model.
Quantum channels, together with an auxiliary ``tensor product'' operator,
can be combined in a network to perform arbitrary computations.
The computational model takes a density matrix as input and produces a density matrix as output.
The entire circuit can be considered as a flow of density matrix transformations.
There is a clear meaning behind this transformation -- namely, a quantum channel.
The specific topology chosen is simply a matter of computational optimization.
For smaller dimensions, the problem can always be optimized directly without assuming any particular topology.
This contrasts with neural networks, where selecting the proper topology is a crucial factor for success.
We approached the problem of reconstructing a quantum channel from data in \cite{belov2024quantumPRE},
but encountered difficulties related to proxy fidelity when going beyond unitary learning.
In this work, these difficulties are mostly overcome.
Formally, the problem is to reconstruct a quantum channel, graphically depicted as a component
\begin{equation}
\hbox{
\begin{circuitikz}[line width=1pt]
\tikzstyle{operator} = [draw,fill=white,minimum size=1.5em] 
\tikzstyle{phase} = [draw,fill,shape=circle,minimum size=5pt,inner sep=0pt]
\tikzstyle{Xcross} = [path picture={ 
\draw[thick,black,inner sep=0pt]
(path picture bounding box.south east) -- (path picture bounding box.north west) (path picture bounding box.south west) -- (path picture bounding box.north east);
}]
\tikzstyle{cross} = [path picture={ 
\draw[thick,black](path picture bounding box.north) -- (path picture bounding box.south) (path picture bounding box.west) -- (path picture bounding box.east);
}]
\tikzstyle{Ocross} = [draw,circle,cross,minimum width=0.3 cm]

\ctikzset{multipoles/thickness=2}
\ctikzset{multipoles/external pins thickness=2}
\ctikzset{multipoles/qfpchip/pin spacing=0.6}

\node [qfpchip,
line width=0.6pt,
num pins=4, no topmark,
external pins width=0.1,
hide numbers,
draw only pins={1,3}
](C){{\Huge $\mathcal{Q}$}};

\node [left, font=\Large] at (C.pin 1) {$\rho$};

\node [right, font=\Large] at (C.pin 3) {$\varrho$};

\end{circuitikz}
}
\label{qcClassicQ}
\end{equation}
converting a density matrix $\rho$ of dimension $n_{\mathcal{Q}}$
to a density matrix  $\varrho$ of dimension $D_{\mathcal{Q}}$
with the transformation (\ref{KrausOperator});
the IN/OUT lines in the circuits correspond not to a qubit,
but to a density matrix of some dimension.
The data is considered classical, and the transformation
(\ref{KrausOperator}) is performed on a classical computer,
possibly hardware-optimized.
In this work, we consider the problem of reconstructing a quantum channel from a classical IN/OUT data sample.

In ML and AI, we typically have $D\le n$,
which means this is a ``summarizing'' quantum channel -- it produces an output state of lower dimension than the input.
One may also consider the opposite situation, where the output dimension is \textsl{greater} than the input.
Such a quantum channel can be considered a ``generative'' one -- it produces a state of higher dimension than the input.
While for unitary learning different input/output dimensions create problems with trace-preservation constraints
(which is why we explored partially unitary learning in \cite{belov2024partiallyPRE}),
quantum channels with a sufficiently high Kraus rank $N_s$ can perfectly
perform mappings between Hilbert spaces of different dimensions.

The paper is organized as follows.
In Section \ref{DataTransform}, we consider the problem of transforming classical vector-to-vector data
into a wavefunction-to-wavefunction (or density-matrix-to-density-matrix) mapping.
This is an important preliminary data-preparation step that enables an effective application of quantum channel transformations.
In most existing work on unitary learning,
authors simply normalize input and output vectors to unit norm and treat them as a form of ``wavefunction''.
We argue that this approach is insufficient, since in most cases it breaks gauge invariance.
Suppose that the input and output vectors are transformed by two non-degenerate linear transformations.
One should require that the resulting quantum channel be independent of the particular transformations chosen.

Then we consider possible criteria of closeness (fidelity) between the ``quantum state''
produced from a given input and the required ``quantum state''.
The choice of fidelity is a broad area of study \cite{wilde2018recoverability, budini2024quantum, johnston2011quantum}.
A key requirement for our approach to be applicable is that the objective function be expressible
as a quadratic function of the quantum channel mapping operators.
For this reason, we restrict ourselves to input data in the form of mappings from mixed states to pure states.
In this setting, the fidelity is exactly quadratic in the mapping operators (without approximation) and,
at the same time,
is sufficient for classical ML/AI input data passed through a quantum channel.
The most general form of the fidelity to which our theory can be applied
is given later in Eq. (\ref{fidelityProjectionsApproximationBBS}) as a ratio of two quadratic forms.
The transition from unitary learning to quantum channel learning allows
us to distinguish between probabilistic mixtures of states and their superpositions.

In Section \ref{Reconst}, we formulate the quantum channel reconstruction problem as a constrained optimization problem.
Two types of constraints are considered: trace preservation and unit-matrix preservation.
The main result of this section is that, when considering a quantum channel of full Kraus rank
$N_s=Dn$,
the resulting optimization problem becomes an exact Semidefinite Programming (SDP) problem.
A remarkable feature of this formulation is that the optimization problem is convex \cite{boyd2004convex}.
This allows the problem to be efficiently solved, for example, using polynomial-time interior-point algorithms.
This is the main reason why we consider a quantum channel as a promising
form of a classical computational model \cite{belov2025superstatePRE}.

In Section \ref{Demonstrations}, we demonstrate an application of SDP software to a number of quantum channel
reconstruction problems. The SDP optimization allows for the reconstruction of quantum channels of different forms,
from unitary to high Kraus rank.
A noticeable feature is that the Kraus rank of the obtained quantum channel typically comprises
less than a few percent of it's maximal possible value. Low Kraus rank has deep significance in AI and ML applications:
our numerical simulations show that a relatively small Kraus rank quantum channel
is typically sufficient to describe experimentally observed classical data.
SDP programming often shows superior results to our eigenvalue-based algorithm discussed in Appendix \ref{RepresentationQC}, especially for high Kraus rank.
A noticeable feature is that, even in unitary learning, the SDP convex optimization approach allows
for the exact reconstruction of the unitary mapping, albeit at the cost of considering a much larger space of Choi matrices.

In Conclusion \ref{conclusion}, we summarize the results and discuss the perspectives
of a quantum channel-based classical computational model arising from the convex form of the
SDP quantum channel reconstruction problem.
An important point is that a single large quantum channel can be split into a network of much smaller ones,
connected in a hierarchical way, known as the density matrix network,
which we introduced in Appendix C of \cite{belov2025superstatePRE}.
This computational model is based on quantum channel transformation (\ref{KrausOperator}),
performed and calculated on a classical computer, possibly hardware-optimized.
The broad field of unitary learning, see \cite{bisio2010optimal,arjovsky2016unitary,hyland2017learning,huang2021learning,yu2023optimal,belov2024partiallyPRE,belov2024quantumPRE}
and thousands of
\href{https://scholar.google.com/scholar?cites=5030720785335451277}{subsequent works},
has become an established field in ML and AI. Using a quantum channel is its natural extension.
The low Kraus rank feature of quantum channels reconstructed from classical data makes them particularly similar.

The interested reader can refer to Ref. \cite{polynomialcode} for the software
implementing our algorithms, build scripts, unit tests, and
integration code for the commercial SDP software
\cite{yamashita2010high,andersen2015cvxopt,diamond2016cvxpy}
used in this paper.

\section{\label{DataTransform}Classical Data Transformation via Quantum Channel}
For classical data, typically in the form of vector-to-vector mapping
\begin{align}
\mathbf{x}^{(l)}&\to\mathbf{f}^{(l)} & l=1\dots M
\label{xfmap}
\end{align}
one needs to construct density matrix entities (in either pure or mixed state)
\begin{align}
\rho^{(l)}  &\to \varrho^{(l)} \label{mixedRhoMap}
\end{align}
that can be passed through the quantum channel (\ref{qcClassicQ}) to obtain the result $\varrho$ from the input $\rho$
with the highest possible fidelity.
See \cite{malyshkin2019radonnikodym, belov2024partiallyPRE, belov2025superstatePRE} for examples of such conversions.
For ML/AI problems, in most cases the mapping takes the form of a pure-state-to-pure-state mapping \cite{belov2025superstatePRE}.
\begin{align}
 \frac{\sum\limits_{j,k=0}^{n-1}x_jG^{\mathbf{x};\,-1}_{jk}x^{(l)}_k}
           {\sqrt{\sum\limits_{j,k=0}^{n-1}x^{(l)}_jG^{\mathbf{x};\,-1}_{jk}x^{(l)}_k}}
  &\to
 \frac{\sum\limits_{j,k=0}^{D-1}f_jG^{\mathbf{f};\,-1}_{jk}f^{(l)}_k}
           {\sqrt{\sum\limits_{j,k=0}^{D-1}f^{(l)}_jG^{\mathbf{f};\,-1}_{jk}f^{(l)}_k}}
  \label{psiXFGflocalizedAppendix}
\end{align}
Here, the $x^{(l)}_k$ and $f^{(l)}_j$ 
are the components of (\ref{xfmap}), and
$x_k$ and $f_j$  
are the components of the arguments of the wavefunctions
\begin{align}
\psi^{(l)}(\mathbf{x})&\to\phi^{(l)}(\mathbf{f})
\label{generalWavefunctionsMapping}
\end{align}
The matrices
$G^{\mathbf{x}}_{jk}=\Braket{x_j|x_k}$ and $G^{\mathbf{f}}_{jk}=\Braket{f_j|f_k}$
are Gram matrices calculated from
$\mathbf{x}^{(l)}$ and $\mathbf{f}^{(l)}$
by averaging over the sample (\ref{xfmap}).
The map in (\ref{psiXFGflocalizedAppendix})
is invariant with respect to arbitrary non-degenerate linear transformations of
$\mathbf{x}$ and $\mathbf{f}$ in the original map (\ref{xfmap}).
If the original map is given as $d\mathbf{x} \to d\mathbf{f}$,
then a similar expression involving the Christoffel function can be used \cite{belov2025tradePRE}.

In this work, we consider sample data in the form of a mapping from a mixed state to a pure state.
\begin{align}
\rho^{(l)}  &\to \Ket{\phi^{(l)}}\Bra{\phi^{(l)}} \label{mixedRhoMapMP}
\end{align}
The reason we limit ourselves to this form is that,
for our theory to be applicable, the total fidelity must be a quadratic function of the mapping operators $B_s$.
The fidelity between a mixed state $\varrho$,
obtained by passing an input state $\rho$ through the quantum channel (\ref{qcClassicQ}),
and a pure state $\Ket{\phi}\Bra{\phi}$, is given by the expected fidelity,
as shown in Eq. (9.89) of Ref. \cite{wilde2011classical}, p.264:
\begin{align}
F(\Ket{\phi}\Bra{\phi},\varrho)&=\Braket{\phi|\varrho|\phi}
\label{fidelityDefinitionPureExpected}
\end{align}
For a general form of mapping (\ref{mixedRhoMap}), the quadratic fidelity is difficult to obtain,
so a proxy (approximation) is required. In \cite{belov2024quantumPRE},
we considered several forms of proxy fidelities and discussed their limitations.
The problem of choosing an appropriate fidelity measure (objective function) is a broad area of
study \cite{wilde2018recoverability, budini2024quantum, johnston2011quantum}.
The key requirement for our approach is that the objective function must be expressible
as a ratio of two quadratic functions of the mapping operators.
We start by considering the weighted fidelity of individual state matches (\ref{fidelityDefinitionPureExpected}).
The weights $\omega^{(l)}$ are typically all equal to 1.
In ML, the weights are sometimes adjusted to account for the importance of specific observations.
The fidelity (\ref{fidelityDefinitionPureExpected}), summed over the sample (\ref{mixedRhoMapMP}),
gives the total fidelity:
\begin{align}
\mathcal{F}&=
\sum\limits_{l=1}^{M} \omega^{(l)}
\sum_{s=0}^{N_s-1} \Braket{\phi^{(l)}|B_s |\rho^{(l)}| B^{\dagger}_s|\phi^{(l)}}
\label{allProjUKxfAppendix}
\end{align}
Expanding the parentheses yields an expression for $S_{jk;j^{\prime}k^{\prime}}$;
Eq. (\ref{SexamplePurePureExact}) is for pure state input $\psi^{(l)}$.
\begin{align}
S_{jk;j^{\prime}k^{\prime}}&=
\sum\limits_{l=1}^{M}\omega^{(l)} \phi^{(l)*}_j\phi^{(l)}_{j^{\prime}}\rho^{(l)*}_{kk^{\prime}}
\label{SexamplePureExact} \\
S_{jk;j^{\prime}k^{\prime}}&=
\sum\limits_{l=1}^{M}\omega^{(l)} \phi^{(l)*}_j\phi^{(l)}_{j^{\prime}} \psi^{(l)*}_{k} \psi^{(l)}_{k^{\prime}}
\label{SexamplePurePureExact}
\end{align}
and the quadratic expression for the total fidelity (\ref{allProjUKxfAppendix})
\begin{align}
\mathcal{F}&=\sum\limits_{s=0}^{N_s-1}\sum\limits_{j,j^{\prime}=0}^{D-1}\sum\limits_{k,k^{\prime}=0}^{n-1}
{B}^*_{jk;s}
S_{jk;j^{\prime}k^{\prime}}
{B}_{j^{\prime}k^{\prime};s}
\label{FidelityBS}
\end{align}
The physical meaning of mixed-to-pure state mapping is essentially that of a reset/replacer-type operation, and it is typically not more useful than pure-state-to-pure-state mapping.
A density matrix is currently seldom used in classical AI/ML applications.
A typical practical problem is the reconstruction of projection operators from normalized pure-state mappings,
which we consider below in Section \ref{projective}.
In terms of practical applications, our work is primarily concerned with pure-state mapping,
since data in this form is currently the only one widely available in ML/AI.
Besides this, we aim to draw attention to the possible benefits of using density matrices for classical ML.
All works we have found that consider density matrices for data encoding
do so only in the context of quantum computers \cite{biamonte2017quantum,schuld2021machine}.
Collecting and analyzing classical data in the form of density matrices could potentially
provide a new perspective on the coherent and non-coherent components of classical data.

In classical ML/AI, the fidelity (\ref{fidelityDefinitionPureExpected}),
which gives the quadratic form (\ref{FidelityBS}),
see also the more general Rayleigh-quotient form (\ref{fidelityProjectionsApproximationBBS}),
may serve as a basis for many approximation methods.
An important feature of (\ref{fidelityDefinitionPureExpected})
is that it provides a quadratic fidelity without increasing the dimensionality of the optimization problem.
In Appendix D of \cite{belov2025superstatePRE}, we showed that Uhlmann fidelity
for mixed-state-to-mixed-state CPTP quantum channel mappings can be represented
as a quadratic function if a number of auxiliary variables (unitary operators $U^{(l)}$) are introduced.
The optimization is then performed simultaneously over the mapping operators ${B}_{jk;s}$
and the auxiliary variables $U^{(l)}$.
While the auxiliary variables allow the fidelity to be formulated
as the required quadratic function of the optimization variables (decision variables),
the resulting increase in dimensionality makes this approach impractical.
The form (\ref{fidelityDefinitionPureExpected})
allows quadratic fidelity to be considered without introducing auxiliary variables,
although it does not apply to general mixed-state-to-mixed-state mappings.

\section{\label{Reconst}Quantum Channel Reconstruction Problem}
The problem of quantum channel reconstruction is reduced to the maximization problem (\ref{FidelityBS}),
subject to constraints on $B_s$.
The most typical constraints correspond to completely positive trace-preserving (CPTP) mappings.
\begin{align}
\delta_{kk^{\prime}}&=\sum\limits_{s=0}^{N_s-1} \sum\limits_{j=0}^{D-1} B_{jk;s}^{*}B_{jk^{\prime};s}
\label{constraintKraussSpurBSKK}
\end{align}
For some ML problems, it is convenient to have an alternative form of constraint where a unit matrix is mapped to a unit matrix.
\begin{align}
\delta_{jj^{\prime}}&=\sum\limits_{s=0}^{N_s-1} \sum\limits_{k=0}^{n-1} B_{jk;s}^{*}B_{j^{\prime}k;s}
\label{constraintKraussSpurBSJJ}
\end{align}
This form arises from partially unitary learning \cite{belov2024partiallyPRE, belov2024quantumPRE}
and corresponds to the condition for a quantum channel (\ref{KrausOperator})
to transform a unit input matrix $A^{IN}$ into a unit output matrix $A^{OUT}$.
This form does not preserve the trace and
is primarily useful for exploring numerical algorithms and studying projections.
This map is completely positive (CP) but not trace-preserving (TP) and should be treated accordingly.
For a unitary mapping, $N_s=1$ and $D=n$,
both forms of constraints coincide.
Other forms of constraints can also be considered.

Additionally, an always-required constraint is that the mapping (\ref{KrausOperator}) must be a completely positive map.
For the Kraus form, this condition is automatically satisfied.

An important feature of optimizing (\ref{FidelityBS}) subject to constraints (\ref{constraintKraussSpurBSKK})
or (\ref{constraintKraussSpurBSJJ}) is that both the objective function and the constraints
are quadratic functions of the mapping operators $B_s$.
This optimization is referred to as a Quadratically Constrained Quadratic Program
(\href{https://en.wikipedia.org/wiki/Quadratically_constrained_quadratic_program}{QCQP}).
When $N_s=1$ and $D=n$, the problem of learning a unitary operator $\mathcal{U}$ from mapping data
is often referred to as ``tomography of unitary operations''
in quantum computation \cite{baldwin2014quantum, holzapfel2015scalable},
or ``unitary learning'' in artificial intelligence \cite{arjovsky2016unitary,belov2024partiallyPRE}.
In \cite{belov2024partiallyPRE} we developed an iterative algorithm that,
in most cases, provides the globally optimal solution to unitary learning. 
We then approached the problem of reconstructing a quantum channel from data in \cite{belov2024quantumPRE},
but encountered difficulties related to proxy fidelity when going beyond unitary learning.
The main difficulties encountered in our previous works were:
\begin{itemize}
\item Although the introduced iterative algorithm was in fact yielding the global maximum solution,
the reason for this behavior was not clear.
\item For maps beyond the unitary case, there was difficulty in formulating a quadratic fidelity
in the mapping operator that satisfies an analog of Uhlmann's theorem.
This was especially problematic for quantum maps that are CP but no longer TP, for example projective operators.
\item Difficulties with algorithm convergence when going beyond Kraus rank-one solutions.
\end{itemize}
In the present work, we re-approach the problem using a different technique: Semidefinite Programming (SDP),
which allows us to reduce the problem to a convex optimization problem.
This explains why we were able to find the global solution.
The admissible forms of fidelity are now extended from a quadratic form (\ref{FidelityBS})
to a ratio of two quadratic forms (\ref{fidelityProjectionsApproximationBBS}).
This extension allows for the reconstruction of a completely new class of operators,
for example projective operators (which are not TP operators).

For numerical optimization the specific method of parametrizing of the space we learn the solution in,
is typically the key to success.
Besides working in the space of Kraus operators ${B}_{s}$, one may also consider an alternative in the form of the Choi matrix $\mathcal{J}$:
\begin{align}
\mathcal{J}(\mathcal{E})&=
\sum\limits_{k,k^{\prime}=0}^{n-1}
\Ket{k}\Bra{k^{\prime}}\otimes \mathcal{E}(\Ket{k}\Bra{k^{\prime}})
\label{ChoiMatrix}
\end{align}
Where the quantum channel mapping $\mathcal{E}$ is the mapping (\ref{KrausOperator}).
The Choi matrix is constructed by mapping the basis states of the system to the corresponding output states under the channel.
Each of the $n\times n$ elements of the input space is mapped to a $D\times D$
output matrix, resulting in a total size of $Dn\times Dn$ elements.
The Choi matrix is quadratic in the Kraus operators $B_s$.
If we express the fidelity (\ref{FidelityBS}) and the constraints (\ref{constraintKraussSpurBSKK})
or (\ref{constraintKraussSpurBSJJ})
in terms of the Choi matrix,
these will become linear functions of its elements.

For the numerical method, it is convenient to define the Choi matrix in the form of swapped indices, specifically:
\begin{align}
\mathcal{J}_{jk;j^{\prime}k^{\prime}}&=
\sum\limits_{s=0}^{N_s-1}
{B}^*_{jk;s}
{B}_{j^{\prime}k^{\prime};s}
\label{Jmatrix}
\end{align}
It has the same dimension $Dn\times Dn$
as the tensor $S_{jk;j^{\prime}k^{\prime}}$ 
and the original Choi matrix (\ref{ChoiMatrix}). The matrix $\mathcal{J}_{jk;j^{\prime}k^{\prime}}$ is Hermitian when considering the 
$jk$ pair as a multi-index $\mathbf{i}=(j,k)$, 
such that $\mathcal{J}_{jk;j^{\prime}k^{\prime}}=\mathcal{J}^*_{j^{\prime}k^{\prime};jk}$;
in numerical implementation, it is convenient to map the multi-index $\mathbf{i}$ to a scalar index as $i=jn+k$.
The quantum channel transformation (\ref{qcClassicQ}) now takes a form similar to the partial trace:
\begin{align}
\varrho_{jj^{\prime}}&=\sum\limits_{k,k^{\prime}=0}^{n-1}\mathcal{J}^*_{jk;j^{\prime}k^{\prime}} \rho_{kk^{\prime}}
\label{KrausOperatorAsChoi}
\end{align}
The objective function (\ref{FidelityBS}) now is:
\begin{align}
\mathcal{F}&=\sum\limits_{j,j^{\prime}=0}^{D-1}\sum\limits_{k,k^{\prime}=0}^{n-1}
\mathcal{J}_{jk;j^{\prime}k^{\prime}}
S_{jk;j^{\prime}k^{\prime}}
\label{FidelityBSJJChoi}
\end{align}
The constraints (\ref{constraintKraussSpurBSKK})
and (\ref{constraintKraussSpurBSJJ}), now take the partial trace forms (\ref{constraintKraussSpurBSKKChoi}) and (\ref{constraintKraussSpurBSJJChoi}) respectively
\begin{align}
\delta_{kk^{\prime}}&=\sum\limits_{j=0}^{D-1} \mathcal{J}_{jk;jk^{\prime}}
\label{constraintKraussSpurBSKKChoi} \\
\delta_{jj^{\prime}}&=\sum\limits_{k=0}^{n-1} \mathcal{J}_{jk;j^{\prime}k}
\label{constraintKraussSpurBSJJChoi}
\end{align}
They are linear functions in the elements of $\mathcal{J}_{jk;j^{\prime}k^{\prime}}$.
In SDP terms, the optimization problem becomes:
Maximize (\ref{FidelityBSJJChoi}) with respect to $\mathcal{J}$, subject to
(\ref{constraintKraussSpurBSKKChoi})
(or (\ref{constraintKraussSpurBSJJChoi}))
constraints \textsl{and} the condition that $\mathcal{J}$ is semidefinite, $\mathcal{J} \succeq 0$.
If one considers the solution without the semidefinite constraint,
imposing only the required linear constraints, the resulting linear-programming (LP) solution produces a matrix $\mathcal{J}$
that generally has both positive and negative eigenvalues, and the obtained value of $\mathcal{F}$
corresponds to the maximal eigenvalue of  $S$.
Whereas in the Kraus representation (\ref{KrausOperator}),
the requirement for a completely positive map is automatically satisfied, for Choi matrix,
an additional constraint of semidefiniteness is always required.
\begin{align}
\mathcal{J} &\succeq 0 \label{SDPconstraint}
\end{align}
There are several possible forms in which to express this requirement.
For example, it could be that $\mathcal{J}$
is required to be the Gram matrix of some vectors:
$\mathcal{J}_{\mathbf{i};\mathbf{i}^{\prime}}=\Braket{v^{[\mathbf{i}]}|v^{[\mathbf{i}^{\prime}]}}$.
Alternatively, one may require $\mathcal{J}$ to have only non-negative eigenvalues.

For the reconstruction of a quantum channel, the problem becomes: maximize the fidelity $\mathcal{F}$
in the form (\ref{SDPobjective})
\begin{align}
\mathrm{Tr} \mathcal{J} S &\to \max \label{SDPobjective} \\
\mathrm{Tr} \mathcal{J} A_c&=\beta_c & c=1\dots N_c \label{SDPconstraintsTr} 
\end{align}
subject to the constraints (\ref{SDPconstraintsTr}) and (\ref{SDPconstraint}).
The optimization is performed with respect to the Choi matrix elements $\mathcal{J}_{jk;j^{\prime}k^{\prime}}$ (\ref{Jmatrix}).
The objective function (\ref{SDPobjective}) is the quadratic fidelity (\ref{FidelityBS}).
The constraints (\ref{SDPconstraintsTr}) are quantum channel mapping constraints,
such as trace preservation (\ref{constraintKraussSpurBSKK})
or possibly some other form such as (\ref{constraintKraussSpurBSJJ}).
There is also an additional constraint requiring complete positivity of the Choi matrix (\ref{SDPconstraint}).
For the more general case in which the fidelity is expressed as a ratio of two quadratic forms,
the objective function and constraints are modified slightly.
See Section \ref{projective} below for this modification, where we study the reconstruction of projective operators.

The number of constraints $N_c$ for (\ref{constraintKraussSpurBSKKChoi}) is $n(n+1)/2$,
and $D(D+1)/2$ for (\ref{constraintKraussSpurBSJJChoi}); the matrices $A_c$ and scalars $\beta_s$ are directly obtained from these.
Without loss of generality, the matrices $A_c$ and $S$ can be taken to be Hermitian.
Also, we will be studying only real symmetric matrices, which are required for ML and AI studies.
In this form, the problem becomes exactly a Semidefinite Programming (SDP) problem
\cite{vandenberghe1996semidefinite,wolkowicz2012handbook,bernard2009moments,anjos2011handbook,wen2013feasible,skrzypczyk2023semidefinite,tavakoli2024semidefinite,taranto2025higher}.
A remarkable feature of this formulation is that the optimization problem is convex \cite{boyd2004convex}.
SDP is defined as a special case of a conic-form convex optimization problem.
Specifically, when the cone is the cone of positive semidefinite matrices (the PSD cone),
the resulting program is convex.
The feasible set in an SDP is the intersection of an affine subspace
(arising from equality constraints or affine linear matrix combinations)
with the cone of positive semidefinite matrices — and this cone is convex.
As a consequence, any local optimum is also a global optimum, and efficient polynomial-time interior-point algorithms exist,
similar to those for LP. This is discussed in \cite{boyd2004convex}.

This explains why identifying a full rank quantum channel is a simpler problem
that unitary learning, which we previously considered \cite{belov2024partiallyPRE}.
The problem of unitary learning is non-convex.
If we consider it as an SDP problem of above, the additional constraint that $\mathcal{J}$
be a rank-one matrix must be imposed, in addition to the requirement $\mathcal{J} \succeq 0$.
The technique developed
in \cite{belov2024partiallyPRE} includes a major transition from traditional mathematical analysis tools
\cite{anjos2011handbook, wen2013feasible, arjovsky2016unitary, hyland2017learning} (e.g., gradient, derivative, etc.)
to using algebraic tools (eigenproblem);
this greatly increases the chances of finding the global maximum.
This technique can also be applied to the quantum channel;
the difficulties we encountered in \cite{belov2024quantumPRE}
were mostly overcome by properly selecting the basis used to represent the quantum channel
(see Appendix \ref{RepresentationQC} below).

However, the problem of full rank quantum channel reconstruction appears to be much simpler -- a convex SDP problem --
as long as the fidelity (\ref{FidelityBS}) takes the quadratic form in the mapping operators,
or more generally, as a ratio of two quadratic forms (\ref{fidelityProjectionsApproximationBBS}).
For this reason, we approach it using interior-point SDP optimization software \cite{yamashita2010high,andersen2015cvxopt,diamond2016cvxpy}.
We also consider it using the updated version of our algorithm \cite{belov2024quantumPRE},
see Appendix \ref{RepresentationQC}.
The main strength of our algorithm is its applicability to non-convex problems.
For example, if we seek a solution as a quantum channel of a specific Kraus rank $N_s$,
the problem becomes non-convex and significantly more difficult to approach.
While our algebraic approach is fully applicable to non-convex problems (such as unitary learning),
it is significantly less computationally efficient than interior-point SDP optimization methods for convex problems.
This difference in computational efficiency becomes especially significant in the case
of high Kraus-rank quantum channels, particularly the full-rank ones.

\section{\label{Demonstrations}Practical Demonstrations of Quantum Channel Reconstruction from Data}
In this section, we examine several quantum channel reconstruction problems employing different methods. We begin with the simplest case of a unitary mapping.

\subsection{\label{UnitaryMappingReconstruction}Reconstructing Unitary Mappings from Data}
Whereas in most practical ML/AI problems the data is given in the form of a general vector-to-vector mapping (\ref{xfmap}),
which must first be converted into a wavefunction mapping such as (\ref{psiXFGflocalizedAppendix}),
unitary dynamics does not require this step, since the data are already provided in wavefunction-mapping form.
Consider a simple classical problem discussed in \cite{belov2024partiallyPRE}.
Let there be an initial state vector $\mathbf{X}^{(0)}$
with unit $\mathrm{L}^2$ norm, and a unitary matrix $\mathcal{U}$.
The operator $\mathcal{U}$ is applied to $\mathbf{X}^{(0)}$ $M$ times:
\begin{align}
\mathbf{X}^{(l+1)}&=\mathcal{U}\mathbf{X}^{(l)}
\label{unitaryDynamcs}
\end{align}
From this sequence, we construct $M$ observation pairs $(\psi^{(l)},\phi^{(l)})$
by selecting the $(l,l+1)$ elements of the sequence and multiplying them by
random phases $\exp(i\xi_l)$ and $\exp(i\zeta_l)$ (or by $\pm 1$ in the real-valued case).
\begin{subequations}
\label{SampleU}
\begin{align}
\Ket{\psi^{(l)}}&=\exp(i\xi_l)\Ket{\mathbf{X}^{(l)}} \label{psiX} \\
\Ket{\phi^{(l)}}&=\exp(i\zeta_l)\Ket{\mathbf{X}^{(l+l)}} \label{phiX}
\end{align}
\end{subequations}
The mapping then coincides exactly with (\ref{mixedRhoMapMP}), with
$\rho^{(l)}=\Ket{\psi^{(l)}}\Bra{\psi^{(l)}}$ and
$\varrho^{(l)}=\Ket{\phi^{(l)}}\Bra{\phi^{(l)}}$.
The random phases render standard regression-type methods inapplicable;
however, the tensor $S_{jk;j^{\prime}k^{\prime}}$ in (\ref{SexamplePurePureExact}) has all random phases canceled.

The sample (\ref{unitaryDynamcs}) should be information-complete \cite{torlai2023quantum}.
For example, if $\mathcal{U}$ corresponds to a Hamiltonian $H$,
\begin{align}
  \mathcal{U}&=
  \exp \left[-i\frac{\tau}{\hbar} H \right] \label{Uquantum}
\end{align}
and  $\mathbf{X}^{(0)}$ is an eigenvector of $H$,
then every step of the mapping (\ref{unitaryDynamcs}) is reduced to multiplying $\mathbf{X}$ by a phase.
In numerical experiments we select $\mathbf{X}^{(0)}$ at random,
so that the unitary dynamics (\ref{unitaryDynamcs}) involves all eigenvectors of the corresponding Hamiltonian.

The tensor $S_{jk;j^{\prime}k^{\prime}}$ (\ref{SexamplePurePureExact}) is obtained from these $M$ observations,
and we use it to reconstruct $\mathcal{U}$.
Whereas in \cite{belov2024partiallyPRE} we sought $\mathcal{U}$ specifically among unitary operators ($N_s=1$),
here we consider the following two cases:
\begin{itemize}
\item A full-rank ($N_s=n^2$) quantum channel (\ref{KrausOperator})
using interior-point SDP optimization software\cite{yamashita2010high,andersen2015cvxopt,diamond2016cvxpy}.
This problem is convex and is expected to be efficiently solvable.
SDP optimization yields the Choi matrix (\ref{Jmatrix}).
When the solution is recovered exactly, this matrix has rank one,
and the unitary operator $\mathcal{U}$ can be obtained from its corresponding eigenvector.
\item A quantum channel (\ref{KrausOperator}) of specific rank $N_s\ge 1$ using the updated version
of our algorithm from Appendix \ref{RepresentationQC}.
This problem is non-convex, as the quantum channel is not full-rank,
but our algebraic approach is typically capable of finding the global maximum.
This optimization problem yields the quantum channel in the $LL^T$ representation (\ref{LowerDiagParametrization}).
When the solution is recovered exactly, the matrix $B$ (\ref{matrixView3}) of dimension $Dn \times N_s$
has a single nonzero column, which corresponds to the desired unitary operator $\mathcal{U}$.
\end{itemize}
The purpose of this section numerical experiments is to assess how well the new technique can handle
the well-established problem of unitary learning, specifically reconstructing unitary channels (Kraus rank one, $N_s=1$)
using a quantum channel with a much higher Kraus rank.

Here we reproduced the unitary dynamics ($N_s=1$, $D=n$) results of Section IV of \cite{belov2024partiallyPRE}
using the following software:
\begin{itemize}
\item \texttt{CVXPY 1.7.5} SDP optimization software \cite{diamond2016cvxpy}
from \href{https://github.com/cvxpy/cvxpy}{www.cvxpy.org}.
It was installed as \texttt{\seqsplit{pip3\ install\ cvxpy}} and then integrated into our Java codebase with a script
\texttt{\seqsplit{com/polytechnik/sdpexternalsolvers/solver\_cvxpy.py}},
which has a \texttt{json} file input/output interface implemented in
\texttt{\seqsplit{com/polytechnik/sdpexternalsolvers/CVXPYsolver.java}};
this requires \href{https://github.com/google/gson}{Gson} library from \texttt{\seqsplit{github.com/google/gson}} to be installed.
\item \texttt{SDPA 7.3.20} SDP optimization software \cite{yamashita2010high}
from \href{https://sourceforge.net/projects/sdpa/}{sdpa.sourceforge.net}.
It was compiled to OS-native code and then integrated into our Java codebase with the
JNI implementation \texttt{\seqsplit{com/polytechnik/sdpexternalsolvers/SDPAsolver.java}}.
\item An updated version of our algebraic approach algorithm,
as described in Appendix \ref{RepresentationQC}, is implemented in \texttt{\seqsplit{com/polytechnik/kgo/QCInverseProblem.java}}.
\end{itemize}
The reconstruction was tested on unitary dynamics data of the form (\ref{unitaryDynamcs}),
so no preliminary conversion, such as (\ref{psiXFGflocalizedAppendix}), is required.
In these tests, we have a known unitary operator $\mathcal{U}$
that generates an information-complete data sample (\ref{SampleU}),
from which the tensor $S_{jk;j^{\prime}k^{\prime}}$ (\ref{SexamplePurePureExact}) is created.
While the algorithm from Appendix \ref{RepresentationQC} finds the unitary operator directly,
the SDP optimization software finds the Choi matrix (\ref{Jmatrix}).
The unitary operator $\mathcal{U}$ is then obtained from $\mathcal{J}_{jk;j^{\prime}k^{\prime}}$
as its eigenvector, corresponding to the maximal eigenvalue.

\begin{figure}[t]

\includegraphics[width=0.9\columnwidth]{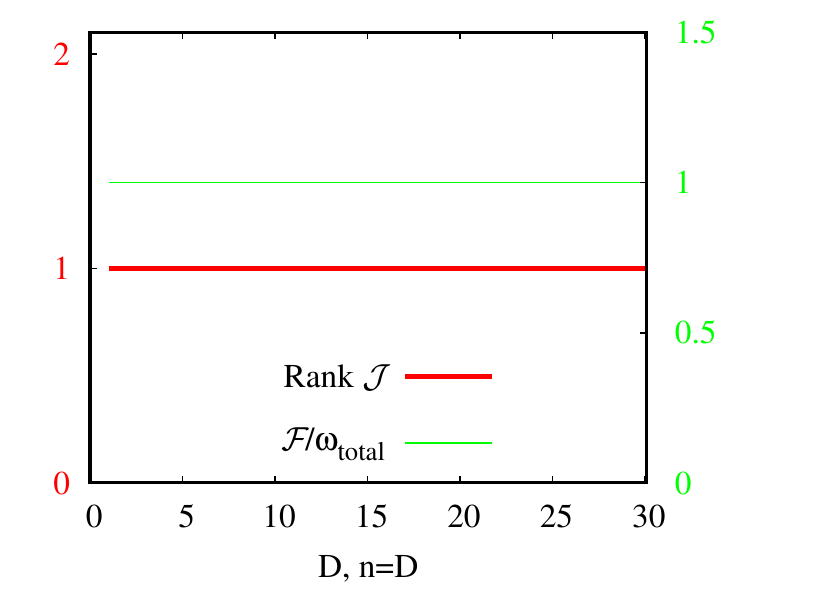} 
  \caption{\label{resRanknUnitaryMapping}
Rank of the Choi matrix $\mathcal{J}$ (\ref{Jmatrix}) (red),
obtained as a solution to the optimization problem (\ref{SDPobjective}) for unitary mapping (\ref{SampleU}) data, is shown.
This plot demonstrates a perfect reconstruction of the unitary operator from the data
for the dimension range $1 \leq D = n \leq 30$. We made similar plots in \cite{belov2024partiallyPRE} using our original software.
The presented plot was generated using the off-the-shelf SDP software: \texttt{CVXPY} and \texttt{SDPA},
both of which reconstruct the unitary channel exactly.
The fidelity (green) of the reconstructed channel is exact because the quantum channel was reconstructed perfectly,
and the original unitary mapping maps a pure state to a pure state (\ref{unitaryDynamcs}),
yielding the exact fidelity (\ref{FidelityBS}).
The purpose of this plot is to demonstrate the application of the convex SDP optimization approach to the problem of unitary learning.
}
\end{figure}

The results for SDP software \texttt{CVXPY} and \texttt{SDPA} are presented in Fig.  \ref{resRanknUnitaryMapping}.
The unitary mapping exact reconstruction results from \cite{belov2024partiallyPRE} were reproduced with the software in question.
The simulation results can be summarized as follows. For small $n$, the algorithm from Appendix \ref{RepresentationQC} is typically faster,
as it works in a space of $\dim_{\mathds{1}}$ (\ref{dimJJ}) parameters,
whereas the SDP algorithms work in a space of the Choi matrix with dimension $n^2\times n^2$.
If one attempts to put an explicit constraint on the rank of the Choi matrix (which is equal to one in the case of unitary learning),
this makes the problem non-convex and not directly solvable with SDP software.
However, the mentioned SDP software was able to find the rank-one Choi matrix exactly for sufficiently large $n$,
without an explicit constraint on the Choi matrix rank.
For a 16-core 5GHz x64 CPU with 32GB of memory,
\texttt{CVXPY} has a maximum $n$ of about 22 (Choi matrix of dimension $484 \times 484$),
and is limited by the calculation time.
For \texttt{SDPA}, the maximum $n$ is about 50 (Choi matrix of dimension $2500 \times 2500$),
and is typically limited by memory.
The algorithm from Appendix \ref{RepresentationQC} is limited by the eigenproblem dimension (\ref{dimJJ}).
Note that most regular commercial software uses \texttt{int32} array indices,
which limits the use of matrices with dimensions larger than $\sqrt{2^{31}}\times\sqrt{2^{31}}$, i.e., $46340 \times 46340$.
For larger matrices, all the libraries must be recompiled with \texttt{int64} array indices.

We conclude that, for unitary dynamics data,
the SDP optimization allows for the exact identification of the unitary dynamics,
despite this being a difficult case of a rank-one Choi matrix.
There was no failure in SDP reconstruction of the thousands of randomly generated unitary matrices we tried,
in addition to those from \cite{belov2024partiallyPRE}.
These numerical experiments show that the convex nature of the SDP optimization problem makes
the difficulties mostly technical, which can be overcome with algorithm improvements and possibly with hardware optimization.

\subsection{\label{QuantumChannelReconstructionToy}Toy Example in Quantum Channel Reconstruction}
Before we consider applications for the practical reconstruction of a quantum channel,
it would be convenient to first examine the quantum channels that allow for achieving 100\%
fidelity in pure state mapping. When the fidelity is exact,
this greatly simplifies the assessment of algorithm correctness and sensitivity to problem degeneracy.

A pure-to-pure state mapping would not give 100\% fidelity for a general CPTP quantum channel,
except for unitary channels and quantum channels with $N_s=1$ and $D \ge n$
that satisfy the trace preserving constraints.

In this section, we consider a CPTP quantum channel with $D>n$ that satisfies (\ref{constraintKraussSpurBSKK}).
This channel has Kraus rank one, preserves the trace, and is not unitary.
To some extent, it is similar to partially unitary channels with $D<n$ satisfying (\ref{constraintKraussSpurBSJJ})
that we considered in \cite{belov2024partiallyPRE}.

These forms have Kraus rank one and computational properties similar to the unitary optimization considered in the previous section.
To construct such a channel, we randomly generated $B_{jk;s}$ operators, now with $D>n$.
These are actually the operators $B_{jk;s}$ from Appendix \ref{RepresentationQC},
but without the lower-diagonal condition (\ref{matrixView3}).
Using the transform (\ref{G05transformKK}), we obtain the $\widetilde{B}_{jk;s}$
operators that satisfy the trace-preservation constraint (\ref{constraintKraussSpurBSKK}).
Then, we randomly generated a wavefunction $\psi^{(l)}$, and calculated $\phi^{(l)}$
by passing it through the quantum channel. The rest is similar to the unitary optimization above:
we constructed the tensor $S_{jk;j^{\prime}k^{\prime}}$ (\ref{SexamplePurePureExact})
and solved the optimization problem using the same software as for the unitary learning above.

The results are the following.
The \texttt{CVXPY} and Appendix \ref{RepresentationQC} software were able to always exactly reconstruct the original
$D>n$
quantum channel used to construct the sample.
The \texttt{SDPA} often misidentifies the problem as being unbounded and frequently fails to find the solution;
this can, to some extent, be resolved by adjusting the parameter \texttt{lambdaStar}.
Overall, the results are very similar to those of the unitary learning case above,
where SDP optimization is limited by the dimension of the Choi matrix,
and the Appendix \ref{RepresentationQC} software is limited by the dimension of the eigenproblem (\ref{dimKK}).
The purpose of this toy-channel example is to present a simple, non-unitary quantum channel that can be reconstructed exactly.

\subsection{\label{QuantumChannelReconstruction}Reconstructing Quantum Channel Mappings from Data}
Whereas unitary learning
is an established field of ML and AI, the application of general quantum channel mapping (\ref{KrausOperator})
to ML and AI represents a novel form of knowledge representation,
allowing one to distinguish between probabilistic mixtures of states and their superposition.
We approached this representation with an eigenproblem-based algorithm back in \cite{belov2024quantumPRE,belov2025superstatePRE},
but some problems still remain, as discussed in Appendix \ref{NumImplementationNotes} below.

The convex nature of the SDP problem offers a new perspective on the direct reconstruction of quantum channel mappings
from experimental data. The only important limiting factor here is the requirement for fidelity
to be a ratio of two quadratic functions of the mapping operators.
When a proxy fidelity (approximation) is used to build a quadratic form,
the maximum of the proxy fidelity may not be achieved for the original quantum channel.
In Table I of \cite{belov2024quantumPRE}, we demonstrated, for a Kraus rank-3 quantum channel,
a situation where a single unitary operator provides a higher proxy fidelity than the proxy fidelity
of the entire rank-3 quantum channel.
For true fidelity, the situation is the opposite: the fidelity of the rank-3 quantum channel is 100\%,
and any unitary operator gives a much lower fidelity.
This makes the selection of the specific form of fidelity a crucial factor in quantum channel reconstruction.

However, this is typically not an issue in the ML and AI fields, where vector-to-vector mapping (\ref{xfmap})
is usually converted to pure state mapping, such as (\ref{psiXFGflocalizedAppendix}),
allowing the use of quadratic fidelity in mapping operators (\ref{fidelityDefinitionPureExpected}).
At the same time, for a pure state-to-pure state mapping, a quantum channel that provides 100\% fidelity
for pure state mapping
may not exist,
which complicates the evaluation of the quality of the solution obtained.
If we find the solution, how can we prove that it is the optimal one?
One simple method is to use several different software packages and compare the results.

A general quantum channel (\ref{KrausOperator})
mapping mixed states (\ref{mixedRhoMap}) requires the fidelity between the obtained $\varrho$ and the desired $\sigma$ density matrices, as described in 
\cite{nielsen2010quantum} p. 409,
\cite{wilde2011classical} p. 285,
\begin{align}
F(\sigma,\varrho)&=\left|\mathrm{Tr}\sqrt{\varrho^{1/2}\sigma\varrho^{1/2}}\right|^2
\label{fidelityDefinitionTextBook}
\end{align}
instead of the familiar expression in Eq. (\ref{fidelityDefinitionPureExpected}),
which gives a quadratic total fidelity (\ref{allProjUKxfAppendix}) in the mapping operators $B_s$.
Preferably, we would like to have the $\sigma, \varrho$ states that allow us to write the fidelity as
\begin{align}
F(\sigma,\varrho)&=\mathrm{Tr}\, \sigma\varrho
\label{fidelityDefinitionTextBookSpur}
\end{align}
in order to obtain the quadratic expression (\ref{FidelityBS}) with the tensor
\begin{align}
S_{jk;j^{\prime}k^{\prime}}&=
\sum\limits_{l=1}^{M}\omega^{(l)} \varrho^{(l)*}_{jj^{\prime}}\rho^{(l)*}_{kk^{\prime}}
\label{SexampleMixedExact}
\end{align}
However, (\ref{fidelityDefinitionTextBookSpur}) matches (\ref{fidelityDefinitionTextBook})
only when one of the density matrices, $\sigma$ or $\varrho$, is a pure state.
This particularly explains why the form in (\ref{fidelityDefinitionPureExpected}) is sufficient in most
ML and AI classical applications, where the desired density matrix $\sigma$ is a pure state,
typically obtained from a vector-to-vector mapping (\ref{xfmap})
in a manner similar to (\ref{psiXFGflocalizedAppendix}).

\subsubsection{\label{ChoiRank}Low Choi Matrix Rank as an Intrinsic Feature of Optimization Solutions}
We start with the task of identifying the Choi matrix structure.
First, a large sample $M\gg Dn$ of wavefunctions (\ref{generalWavefunctionsMapping})
was randomly generated for various values of $D$ and $n$.
The case $D<n$ is typically more important in practical applications, for example as a feature-identification task.
After the computation of the tensor $S_{jk;j^{\prime}k^{\prime}}$ (\ref{SexampleMixedExact}), the optimization problem is solved.
The matrix $S_{jk;j^{\prime}k^{\prime}}$
generated in the described way is non-degenerate, with eigenvalues approximately more or less evenly distributed within some range.
Effectively, this is a random matrix obtained as a fourth-degree correlator of two vectors $\psi$ and $\phi$ (\ref{SexamplePurePureExact}),
with the vectors now chosen randomly.

The optimization is approached with the SDP algorithm to solve the optimization problem (\ref{SDPobjective})
subject to the TP constraints (\ref{constraintKraussSpurBSKKChoi}) and CP constraints (\ref{SDPconstraint}).
The solution results in the Choi matrix (\ref{Jmatrix}) being obtained.
The \texttt{CVXPY} and \texttt{SDPA} software perform quite well for this task;
\texttt{SDPA} is faster and gives better precision, but sometimes misidentifies the problem as being unbounded.
Our Appendix \ref{RepresentationQC} software has some other issues, which are discussed below in Appendix \ref{NumImplementationNotes}.

Having these tools available, the first question to ask is about the rank of the obtained Choi matrix.
When fitting random data using traditional learning representations,
such as least squares, PCA, neural networks, and others,
for random input/output data,
the information typically propagates to all available degrees of freedom of the model.
Numerical experiments show that this does not seem to be the case for quantum channel mapping:
The obtained Choi matrix typically has a rank much lower than its maximal possible rank $Dn$.
To identify the Choi matrix structure, we performed the following numerical experiment:

For a given  $n$ and $D$, we randomly generate $M=2n^2D^2+100000$ normalized wavefunction pairs as pseudo-random unit-norm vectors of length
$n$ ($\psi^{(l)}$)
and $D$ ($\phi^{(l)}$),
considering them as if they were the mapping (\ref{generalWavefunctionsMapping}).
Here and below, the phrase ``generated random matrix/vector'' refers to using a $[-1:1]$
uniformly distributed pseudo-random generator, \texttt{\seqsplit{java.util.Random}},
with a specific initial seed chosen to ensure reproducibility of the numerical experiments.
We used 20 different initial seeds to compare the results.
Although the resulting distributions were similar across runs, using different seeds allowed us to confirm the observed effects.
From the obtained mapping, we construct the tensor  $S_{jk;j^{\prime}k^{\prime}}$ (\ref{SexamplePurePureExact})
and solve the optimization problem with it.
The result of the optimization yields the Choi matrix (\ref{Jmatrix}) of dimension $Dn\times Dn$.
All the eigenvalues of the Choi matrix are obtained.
Many of them are small, making it difficult to distinguish actually small values from those due to numerical errors.
For this reason, we define the rank of the Choi matrix as follows: for sorted eigenvalues, going down from the largest ($i=Dn-1$)
to the smallest ($i=0$), we stop when either the eigenvalue is lower than  $10^{-5}$
or the ratio to the previous eigenvalue $\lambda^{[i+1]}/\lambda^{[i]}$
 exceeds $10^4$. 
\begin{figure}[t]

\includegraphics[width=0.9\columnwidth]{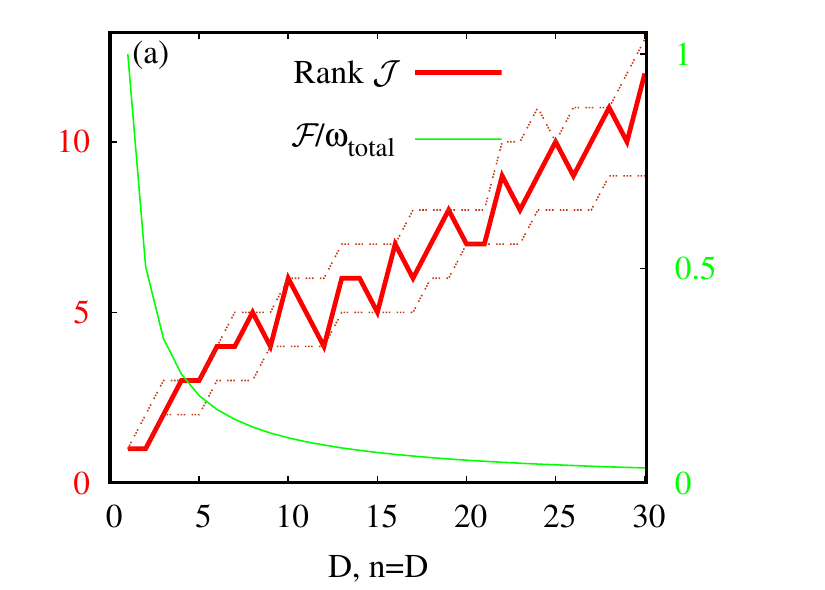} \\
\includegraphics[width=0.9\columnwidth]{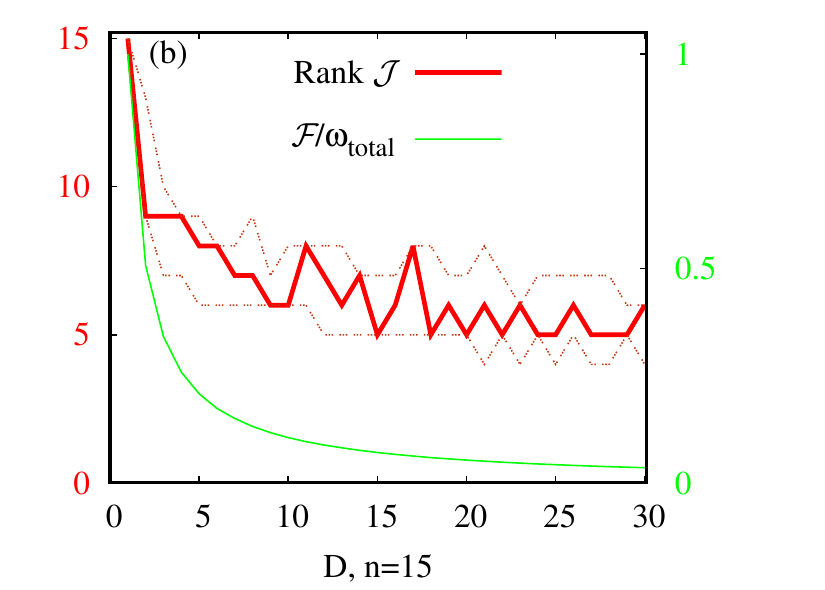}
  \caption{\label{resRankn}
The rank of the Choi matrix $\mathcal{J}$ (\ref{Jmatrix}) (red)
obtained as a solution to the optimization problem for a random
 $\psi\to\phi$ mapping (\ref{generalWavefunctionsMapping}):
(a) The case $1\le D=n \le 30$.
The rank grows slowly with the increase of $n$;
the rank value typically comprises less than 1.5\% of the matrix dimension $Dn$.
(b) The case where $n=15$ is fixed and $1\le D\le 30$ is varied.
The maximal obtained Choi matrix rank is $15$ for $D=1$ (trace calculation);
the rank decreases slowly with an increase in $D$.
The dashed lines correspond to the minimum and maximum ranks obtained across 20 runs with different initial seeds,
while the solid line corresponds to a particular seed chosen as a representative example.
Further increasing the number of runs resulted in little to no change in the observed minimum and maximum rank values.
The green line represents the relative fidelity $\mathcal{F}/\sum_l \omega^{(l)}$;
error bars for the fidelity are not shown because the variation across the 20 runs is very small.
Since the sample is random, the fidelity decreases with the increase in problem size,
as it becomes less likely for two random states to match as the dimension increases.
For $D=1$, the fidelity is exact: $\mathcal{F}=\sum_l \omega^{(l)}$; this is simply a trace calculation.
These plots show that it is difficult to obtain a high-rank Choi matrix as
the solution to the quantum channel reconstruction optimization problem.
Quantum channels reconstructed from classical data usually exhibit a low Kraus rank.
  }
\end{figure}

In Fig. \ref{resRankn}, the rank of the obtained Choi matrix as the solution for
optimization problem with this $S_{jk;j^{\prime}k^{\prime}}$,
corresponding to a completely random mapping (not corresponding to an actual quantum channel),
is presented for two cases: equal IN/OUT dimensions  $1\le D=n\le 30$,
and for fixed $n=15$ 
with varied $1\le D\le 30$.
The optimization problem was solved by \texttt{CVXPY} and \texttt{SDPA} programs.
They typically produce almost identical solutions with identical ranks, with only a few cases of noticeable differences.
As expected, the Choi matrix rank almost does not depend on the initial random value
used to generate random mapping (\ref{generalWavefunctionsMapping}) --
only in a few points the Choi matrix rank may possibly differ slightly.
The dashed lines correspond to the minimum and maximum ranks obtained across 20 runs with different initial seeds.
Further increasing the number of runs resulted in little to no change in the observed minimum and maximum rank values.

The most unexpected result is that the rank of the Choi matrix, corresponding to the optimization problem solution with
$S_{jk;j^{\prime}k^{\prime}}$, which is obtained as a four-particle correlation
$\Braket{\phi^*_j\phi_{j^{\prime}}\psi^*_k\psi_{k^{\prime}}}$ (\ref{SexamplePurePureExact}) of random wavefunction, is small.
For example, Fig. \ref{resRankn}a shows that the rank of the Choi matrix does not exceed $13$ for $D=n=30$,
i.e. it is less than 1.5\% of the matrix dimension.
This is quite unusual. We compare this case with a general case of a random $B_{jk;s}$:
we generated an initial random matrix $B_{jk;s}$ of $N_s=Dn$,
converted it to trace-preserving form (\ref{constraintKraussSpurBSKK}) with the transform (\ref{G05transformKK}),
then obtained the corresponding Choi matrix (\ref{Jmatrix}), and finally determined its rank.
Whereas for the Choi matrix obtained as a solution to the optimization problem, the rank is typically small,
the rank of the Choi matrix obtained from this randomly generated $B_{jk;s}$
is very close to the maximal possible rank  $Dn$, typically equal to or differing by no more than $2-3$ from the maximal value.

In Fig. \ref{resRankn}b, we fixed $n=15$ and varied $1\le D\le 30$.
At $D=1$, the rank is maximal,  $15$,
which corresponds to a quantum channel performing a trace calculation,
as in Eq. (\ref{KrausOperator}), with $B_s$ being an arbitrary orthogonal basis.
As $D$ increases, the rank of the Choi matrix decreases. The fidelity (green line) also decreases with the increase in problem size,
as it becomes less likely for two random states to match as the dimension increases.

These numerical experiments show that quantum channels reconstructed from classical data typically exhibit a low Kraus rank.
This is an important feature to consider when using quantum channels as ML/AI knowledge representations.
It will be shown below that this result of a low-rank Choi matrix is common and occurs not only for randomly generated mappings.
The random mapping is the most representative: if it were a linear regression or PCA-type expansion, it would expand to all available degrees of freedom.
A question arises as to whether the low Kraus rank (about a few percent of the matrix dimension) of the solution $\mathcal{J}$
to the optimization problem (\ref{SDPobjective}) is a specific feature of
the $S_{jk;j^{\prime}k^{\prime}}$ form obtained from ``classical'' data,
as four-particle correlation in (\ref{SexamplePurePureExact}), or if it is an intrinsic property of the optimization problem itself.

\begin{figure}[t]

\includegraphics[width=0.9\columnwidth]{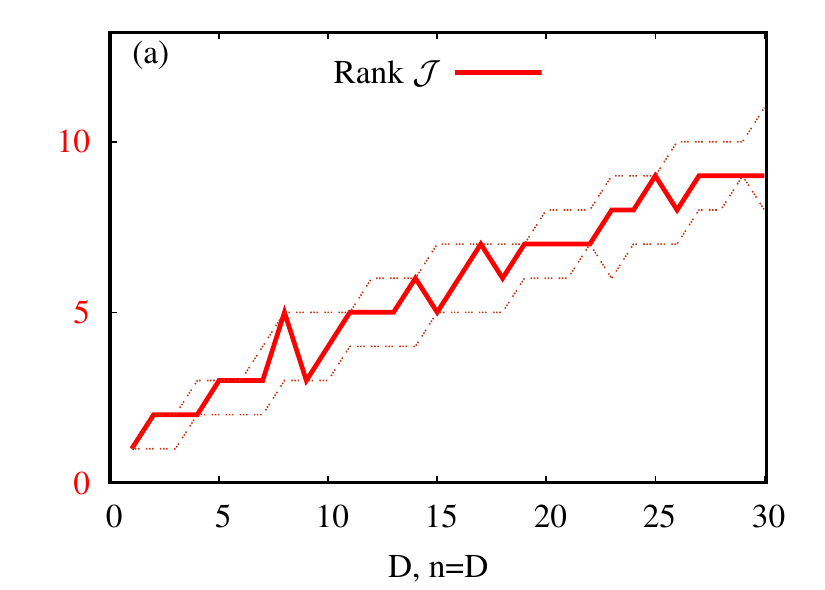} \\
\includegraphics[width=0.9\columnwidth]{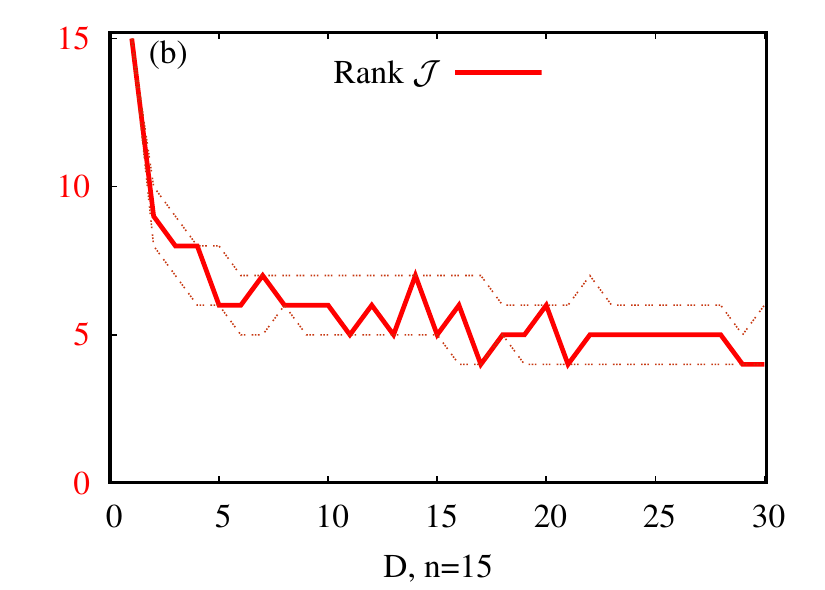}
  \caption{\label{randomMatrixResult}
Rank of the Choi matrix $\mathcal{J}_{\mathbf{i};\mathbf{i}^{\prime}}$ (\ref{Jmatrix})
obtained as a solution to the optimization problem for a case with random $S_{\mathbf{i};\mathbf{i}^{\prime}}$,
without any internal structure, is shown.
The dashed lines correspond to the minimum and maximum ranks obtained across 20 runs with different initial seeds,
while the solid line corresponds to a particular seed chosen as a representative example.
Similarly to Fig. \ref{resRankn},
we also obtain a low-rank Choi matrix $\mathcal{J}$.
This leads us to conclude that the low rank of the solution (a few percent of the matrix size $Dn$)
is an intrinsic property of the optimization problem (\ref{SDPobjective}).
  }
\end{figure}

The simplest test would be to consider $S_{\mathbf{i};\mathbf{i}}$ as a random Hermitian matrix with multi-index $\mathbf{i}=(j,k)$,
without any relation to four-particle correlation as in (\ref{SexamplePurePureExact}) or (\ref{SexampleMixedExact}),
and solve the optimization problem with it as if it were the tensor $S_{jk;j^{\prime}k^{\prime}}$.
The result for random matrix in Fig. \ref{randomMatrixResult} is very similar to what we observed for the random sample in Fig. \ref{resRankn}.
Both \texttt{CVXPY} and \texttt{SDPA} produce identical solutions. This fact -- that a random $S_{\mathbf{i};\mathbf{i}}$,
without any internal structure, produces a solution to optimization problem (\ref{SDPobjective})
with a low-rank Choi matrix $\mathcal{J}$ --
leads us to conclude that the low rank of the solution is an intrinsic property of the optimization problem (\ref{SDPobjective}),
and is not related to the way the sample is constructed or how $S_{jk;j^{\prime}k^{\prime}}$ is calculated (\ref{SexampleMixedExact}).

Additionally, we tried a variety of other forms of $S_{jk;j^{\prime}k^{\prime}}$
and have never been able to obtain a Choi matrix solution to (\ref{SDPobjective}) with a rank greater than $\max(D,n)$.
The fact that, for quantum channel reconstruction, the optimization problem solution typically produces a low-rank Choi matrix
is an important property in practice.
Low Kraus rank has deep significance in AI and ML applications:
our numerical simulations show that
a relatively small Kraus rank quantum channel
is typically sufficient to describe experimentally observed data.
This allows the use of a wide range of problem optimizations, particularly hardware-type optimizations.
Most importantly, quantum channel learning representation can be considered a natural extension to unitary learning;
the convex nature of the optimization problem makes the approach especially appealing in practical applications.

\subsubsection{\label{QCFromDataChoiRank}Data-Driven Quantum Channel Reconstruction}

After establishing the important property of low Choi rank in the optimization problem solution,
we test the approach on some actual data.
The main difficulty is that there is no quantum channel (except for unitary and unitary-like channels)
that will map pure states (\ref{generalWavefunctionsMapping}) exactly.
For this reason, for a general quantum channel, we construct a sample that transforms an input pure state $\psi^{(l)}$
to an output mixed state $\varrho^{(l)}$ by passing $\psi^{(l)}$ through the channel, and then use the eigenvector of the output state
$\varrho^{(l)}$
corresponding to the maximal eigenvalue as the resulting pure state $\phi^{(l)}$.
In this way, the quantum channel reconstruction problem can be approached directly,
but the original quantum channel can serve only as a lower-bound estimate for the solution.
Specifically:
\begin{itemize}
\item
We construct full Kraus rank ($N_s=Dn$) operators $B_{jk;s}$
as random matrices with $D,n=1\dots 30$.
This is a completely positive but not necessarily trace-preserving map from a space of dimension  $n$
to a space of dimension $D$.
Using the transform (\ref{G05transformKK}), we obtain the $\widetilde{B}_{jk;s}$
operators that satisfy the trace-preserving constraint (\ref{constraintKraussSpurBSKK}); these are actually the operators
$B_{jk;s}$ from Appendix \ref{RepresentationQC}, but without the lower diagonal condition (\ref{matrixView3}).
As discussed above, the Choi matrix corresponding to this randomly generated quantum channel has
a Kraus rank equal to or very close to the maximal value, $Dn$.

\item For $l=1\dots M$, where $M$ is chosen as $M=2n^2D^2+100000$,
we randomly create wavefunctions $\psi^{(l)}$ of dimension $n$. The state $\Ket{\psi^{(l)}}\Bra{\psi^{(l)}}$
is then passed through the quantum channel (\ref{KrausOperator}), and the density matrix $\varrho^{(l)}$ of dimension $D$ is obtained.
The mixed-state output $\varrho^{(l)}$, however, does not allow the quadratic fidelity (\ref{fidelityDefinitionPureExpected})
to be applied. To obtain the mapping of the form (\ref{mixedRhoMapMP}), we find all eigenvectors of $\varrho^{(l)}$
and select $\Ket{\phi^{(l)}}$ as the one corresponding to the maximal eigenvalue of $\varrho^{(l)}$.
(Note that for a large $M$, this construction of a sample can be slow, but reducing the sample size $M$ does not change the result).
This way, the pure state-to-pure state mapping with quadratic fidelity (\ref{allProjUKxfAppendix}) is obtained,
and the tensor $S_{jk;j^{\prime}k^{\prime}}$ (\ref{SexamplePurePureExact}) is calculated.
This tensor, however, does not correspond to the quantum channel used to construct the sample,
as the sample was modified to fit the pure state-to-pure state paradigm, which is typical in the data analysis field.
Thus, we cannot require the reconstructed channel to match the original quantum channel exactly,
but we can compare the fidelity obtained, using the original quantum channel as a lower bound.

\end{itemize}
This analysis was conducted using \texttt{CVXPY} and \texttt{SDPA} software, which produce identical results.
The difficulties with our eigenproblem-based approach are discussed in Appendix \ref{NumImplementationNotes} below.
The result is presented in Fig. \ref{resRanknForQC} in the same form as in Fig. \ref{resRankn} for the random sample.
The $\mathcal{F}_{init}$ (green) is the fidelity obtained for the pure state-to-pure state sample,
with the quantum channel initially used to construct the sample by taking the eigenvector corresponding
to the highest eigenvalue of the output density matrix $\varrho^{(l)}$.
It decreases with the dimension, but slower than in Fig. \ref{resRankn} for the random sample.
The $\mathcal{F}$ (olive) is the fidelity corresponding to
the quantum channel obtained as the solution to the optimization problem (\ref{SDPobjective}) with the sample we created.
At high dimensions, it exceeds $\mathcal{F}_{init}$ by several times.
This simulation confirms an important feature of the low Kraus rank of the resulting Choi matrix.

\begin{figure}[t]

\includegraphics[width=0.9\columnwidth]{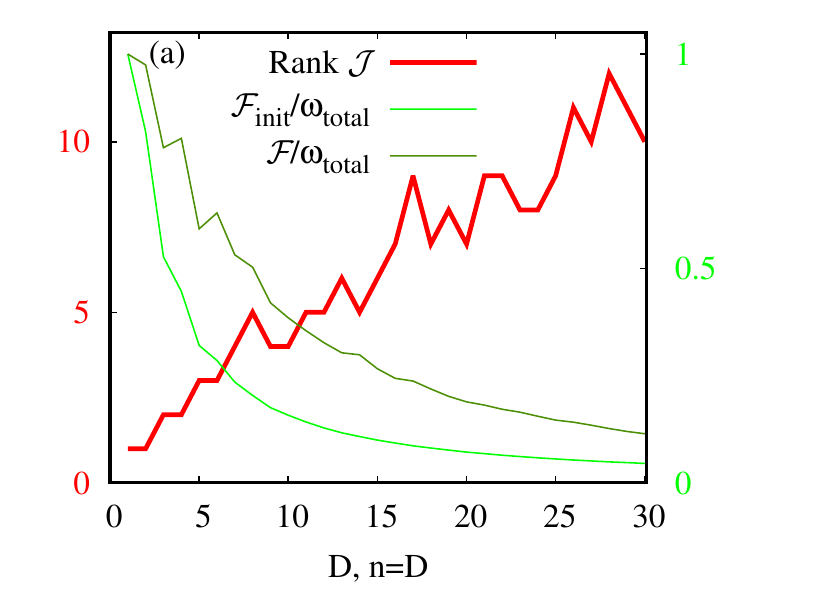} \\
\includegraphics[width=0.9\columnwidth]{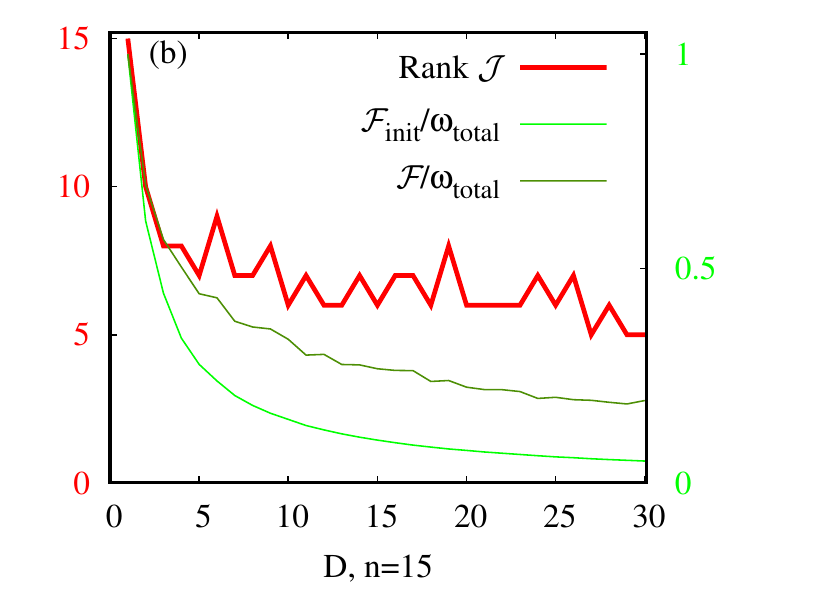}
  \caption{\label{resRanknForQC}
Rank of the Choi matrix $\mathcal{J}$ (\ref{Jmatrix}) (red)
obtained as a solution to the optimization problem for a sample calculated as
a random pure state passed through the quantum channel,
with the eigenvector corresponding to the maximal eigenvalue of the output density matrix taken as the result.
The fidelity $\mathcal{F}_{init}/\sum_l \omega^{(l)}$ (green) is the fidelity of this initial map on the generated sample.
The fidelity $\mathcal{F}/\sum_l \omega^{(l)}$ (olive) is the fidelity of the quantum channel obtained
as a result of the optimization problem (\ref{SDPobjective}) with $S_{jk;j^{\prime}k^{\prime}}$ (\ref{SexamplePurePureExact}) from this sample.
At high dimensions, it is several times higher than the fidelity of the original map.
The low Kraus rank feature is also confirmed for this sample.
(a) The case $1\le D=n \le 30$.
(b) The case where $n = 15$ is fixed and $1\le D\le 30$ is varied.
When $D=1$, trace preservation reduces to the trace-calculating quantum channel, which has Kraus rank $N_s=n$.
The error bars are not shown because the plot already contains a large number of curves.
They are very similar to those in Fig. \ref{resRankn} across all 20 runs.
  }
\end{figure}

Whereas we cannot theoretically confirm that the found solution is exactly optimal,
the identical results obtained from \texttt{CVXPY} and \texttt{SDPA} software,
along with the convexity property of the SDP problem itself, lead us to conclude that the practically found fidelity maximum is global.
These simulations confirm the applicability of the SDP technique to quantum channel
reconstruction and show that a quantum channel with a relatively small Kraus rank
is typically sufficient to describe experimentally observed data.

\subsection{\label{InverseQuantumChannel}Approximate Inverse of a Quantum Channel}
Consider a CPTP map (\ref{KrausOperator}) subject to the TP constraints (\ref{constraintKraussSpurBSKK}),
and ask under what conditions this channel can be inverted.
An invertible quantum channel is one whose action can be undone by another CPTP map. This requirement is very restrictive. A quantum channel is invertible (i.e., admits a CPTP inverse) if and only if it is a unitary channel.
This corresponds to physical reversibility with no loss of information.
In all other cases, one can obtain only one of the following weaker notions of invertibility:
\begin{itemize}
\item A linear inverse, which may exist as a linear map but is not necessarily completely positive.
\item An inverse on a restricted set of states, for example, one defined on a known subspace.
\item An inverse with post-selection, which may be linear but is not necessarily trace-preserving.
\item An approximate inverse, in which some irreversibility error remains.
\end{itemize}
It is important to note that for pure state-to-pure state mapping, the fidelity in the form (\ref{fidelityDefinitionPureExpected})
gives exactly the same $S_{jk;j^{\prime}k^{\prime}}$ (\ref{SexamplePurePureExact}) for both the direct and inverse quantum channels,
with the indices swapped. Note that the trace-preserving constraints (\ref{constraintKraussSpurBSKK})
are the same as the unit matrix invariance constraint (\ref{constraintKraussSpurBSJJ}), with the indices swapped.
Thus, the maximization of the fidelity (\ref{FidelityBS}) subject to the (\ref{constraintKraussSpurBSKK})
constraint gives the direct quantum channel, and the maximization of the same fidelity subject to (\ref{constraintKraussSpurBSJJ})
gives the inverse quantum channel. This follows immediately from swapping $A^{IN}/A^{OUT}$ in (\ref{KrausOperator})
and applying the trace-preservation constraints to them;
similarly, in (\ref{KrausOperatorAsChoi}) for $\rho\leftrightarrow\varrho$.
Whereas for small $N_s$ the constraints cannot always be satisfied, for large enough Kraus rank,
the quantum channel satisfies all the required constraints.

The problem of constructing an inverse quantum channel can be an attractive approach in certain ML/AI applications.
It allows one to consider mappings from pure states to mixed states, a common setting in quantum information processing.
In this case, the mapping becomes a quadratic function of the mapping operators,
in the same way that the mapping (\ref{mixedRhoMapMP}) from a mixed state to a pure state yields
a quadratic fidelity in the mapping operators for the direct quantum channel.

Our numerical experiments show that the inverted quantum channel provides similar properties to the direct one in terms
of Choi matrix rank. A question arises: when can the inverse channel in ML applications be constructed simply by
``swapping the indices'', besides for the unitary channel?
If we drop the requirement of trace preservation, and require a single ``simplified'' condition (\ref{L2norm})
-- this allows the fidelity optimization problem to be reduced to an eigenvalue problem.
The question of constructing alternative forms of constraints (not necessarily trace-preserving)
that would allow for the direct construction of an inverse quantum channel is a subject of future research.
In data analysis, other forms of constraints beyond trace preservation are often crucial.

\subsection{\label{projective}Reconstructing Projective Operators: Approaches and Applications}
Besides constructing the inverse quantum channel (in this case, we typically have $D=n$),
the unit matrix-to-unit matrix constraints (\ref{constraintKraussSpurBSJJ}) in the $D<n$
case arise in the reconstruction of projections.
Reconstructing projective operators is of particular interest in data analysis.
Consider a space of dimension $n$, and a projective operator $P$, projecting it to a smaller subspace of dimension $D$.
Instead of the common condition $P^2=P$, it is better to represent the unit matrix in the original space as
\begin{align}
\mathds{1}&=\sum\limits_{k=0}^{n-1}\Ket{\psi^{[k]}}\Bra{\psi^{[k]}}
\label{UnitMatrixInBasis}
\end{align}
where  $\psi^{[k]}$ is an arbitrary orthogonal basis in the input vector space.
Then we can select this basis as the direct sum of the desired subspace of dimension $D$
and the orthogonal subspace, which has dimension $n-D$.
From here, we immediately obtain that a projective operator is an operator that transforms a unit matrix of dimension
$n$ into a unit matrix of dimension  $D$,
That is, it is a Kraus rank-one ($N_s=1$) quantum channel with  $D<n$,
satisfying the constraints (\ref{constraintKraussSpurBSJJ}). This is a trace-decreasing map.

Assume we have a sample (\ref{generalWavefunctionsMapping}) with the mapping:
\begin{align}
\psi^{(l)} &\to
\phi^{(l)}=\frac{\Ket{P|\psi^{(l)}}}{\sqrt{\Braket{\psi^{(l)}|P^{\dagger}|P|\psi^{(l)}}}}
\label{mappingProjPsiPhi}
\end{align}
where $P$ is a projective operator. This means that $\phi^{(l)}$
is a normalized projection of $\psi^{(l)}$ onto some subspace.
A trivial way of using the direct projections  $\Braket{\psi|\phi}$ to reconstruct $P$
is usually inapplicable, since random phases, similar to those in (\ref{SampleU}),
are always assumed to exist in the measured data (these phases are $\pm 1$ for real-valued data).
The problem is reduced to maximizing the fidelity
\begin{align}
\mathcal{F}&=\sum\limits_{l=1}^{M} \omega^{(l)}\frac{\left|\Braket{\phi^{(l)}|\mathcal{U}|\psi^{(l)}}\right|^2}
{
\Braket{\psi^{(l)}|\mathcal{U}^{\dagger}|\mathcal{U}|\psi^{(l)}}
} \xrightarrow[\mathcal{U}]{\quad }\max
\label{fidelityProjections}
\end{align}
for partially unitary operators $\mathcal{U}$ with  $D<n$ that satisfy the constraints (\ref{constraintKraussSpurBSJJ}).
These constraints are quadratic in $\mathcal{U}$ and are perfectly applicable to the framework of this paper.
However, the fidelity in (\ref{fidelityProjections}) cannot be exactly reduced to the quadratic form (\ref{FidelityBS}),
since for each $l$, there is an $\mathcal{U}$-dependent term in the denominator.

One can formally approximate (\ref{fidelityProjections}) not by summing the individual terms,
but instead by summing the numerator and denominator terms separately, and then considering their ratio:
\begin{align}
\mathcal{F}&\approx
\frac{
\sum\limits_{l=1}^{M} \omega^{(l)}\left|\Braket{\phi^{(l)}|\mathcal{U}|\psi^{(l)}}\right|^2
}
{
\sum\limits_{l=1}^{M} \nu^{(l)}
\Braket{\psi^{(l)}|\mathcal{U}^{\dagger}|\mathcal{U}|\psi^{(l)}}
} \xrightarrow[\mathcal{U}]{\quad }\max
\label{fidelityProjectionsApproximation}
\end{align}
where $\nu^{(l)}$ are some weights.
One may choose for example $\nu^{(l)}=\omega^{(l)}$ and thus normalize (\ref{fidelityProjectionsApproximation})
to the $[0:1]$ interval.
As discussed in \cite{belov2024quantumPRE}, an important property required for any approximated fidelity
is that it should attain its maximum when the mapping operators correspond to the sought quantum channel.
For counterexamples, see Table I of \cite{belov2024quantumPRE}.
Our numerical experiments show that for the fidelity (\ref{fidelityProjectionsApproximation}),
when choosing $\nu^{(l)}=\omega^{(l)}$,
the maximum corresponds to the projective operator used to construct the sample,
i.e., the projective operator is recovered exactly from the sample in (\ref{mappingProjPsiPhi}).

This approximate expression represents the total fidelity as the ratio of two quadratic forms in the mapping operators.
Whereas the total fidelity forms we previously considered were always
a weighted sum of individual state matches (\ref{allProjUKxfAppendix}),
the total fidelity in the form of (\ref{fidelityProjectionsApproximation}) differs from all of these:
it aggregates the numerator and denominator separately and then considers their ratio.
Numerical experiments show that this form of fidelity allows for the exact reconstruction
of the projective operator.
This is a critically important feature. See Table I of \cite{belov2024quantumPRE},
where we demonstrated that for many fidelity approximations,
their maximum is not reached for the channel used to create the mapping.
For a quantum channel with $N_s\ge 1$,
the fidelity takes the following form, which we refer to as a Rayleigh-quotient-type fidelity:
\begin{align}
\mathcal{F}&=
\frac{
\sum\limits_{s=0}^{N_s-1}\sum\limits_{j,j^{\prime}=0}^{D-1}\sum\limits_{k,k^{\prime}=0}^{n-1}
{B}^*_{jk;s}
S_{jk;j^{\prime}k^{\prime}}
{B}_{j^{\prime}k^{\prime};s}
}
{
\sum\limits_{s=0}^{N_s-1}\sum\limits_{j,j^{\prime}=0}^{D-1}\sum\limits_{k,k^{\prime}=0}^{n-1}
{B}^*_{jk;s}
Q_{jk;j^{\prime}k^{\prime}}
{B}_{j^{\prime}k^{\prime};s}
}
 \xrightarrow[B]{\quad }\max
\label{fidelityProjectionsApproximationBBS}
\end{align}
where $S_{jk;j^{\prime}k^{\prime}}$ is the familiar tensor from (\ref{SexamplePurePureExact}) and 
$Q_{jk;j^{\prime}k^{\prime}}$ is a similar tensor obtained in the same way from the denominator of (\ref{fidelityProjectionsApproximation})
\begin{align}
Q_{jk;j^{\prime}k^{\prime}}&=
\delta_{jj^{\prime}}\sum\limits_{l=1}^{M}\nu^{(l)} \psi^{(l)*}_{k} \psi^{(l)}_{k^{\prime}}
\label{QexamplePurePureExact}
\end{align}
Note that even if the basis $\psi_k$
is chosen such that $\delta_{kk^{\prime}}=\Braket{\psi_{k}|\psi_{k^{\prime}}}$ holds for the actual relation,
the sum in (\ref{QexamplePurePureExact}) is not equal to $\delta_{kk^{\prime}}$.
This sum is over the sample of input vectors in (\ref{mappingProjPsiPhi});
this sample typically consists of a number of unrelated observations (e.g., a collection of images)
mapped to some of its features (e.g., image content property).
The sum $\sum_{l=1}^{M}\nu^{(l)} \psi^{(l)*}_{k} \psi^{(l)}_{k^{\prime}}$
characterizes the learning sample, not the relationship between $\psi$ and $\phi$.
In this sense, the tensor $S_{jk;j^{\prime}k^{\prime}}$ (\ref{SexamplePurePureExact})
contains both: the information about the learning sample and the relation (\ref{mappingProjPsiPhi}) that we want to learn.
We discussed above the necessity for the initial conversion of the learning sample to
a form that can be treated as a wavefunction mapping.
Eq. (\ref{psiXFGflocalizedAppendix}) is the simplest such form.

The expression in (\ref{fidelityProjectionsApproximationBBS}) is likely the most general form of the fidelity to which our theory can be applied.
Whereas most of the existing works use a fidelity (such as in (\ref{fidelityDefinitionTextBook}))
between individual states and then derive the total fidelity of a sample
(such as the sample (\ref{mixedRhoMapMP}) with fidelity (\ref{allProjUKxfAppendix})),
the form in (\ref{fidelityProjectionsApproximationBBS}) defines the total fidelity directly,
without the intermediate step of state matching.
Numerical experiments show that this form allows for the exact reconstruction of projective operators
from the pure state mapping data sample (\ref{mappingProjPsiPhi}).

The algorithm from Appendix \ref{RepresentationQC} is directly applicable
(at least in the $N_s=1$ case of finding regular projective operators) by replacing the simplified constraint in (\ref{L2norm})
with the denominator of (\ref{fidelityProjectionsApproximationBBS}).

The problem can also be reduced, for an arbitrary $N_s$, to an SDP optimization problem using a similar technique.
The constraints, such as (\ref{constraintKraussSpurBSKK}) or (\ref{constraintKraussSpurBSJJ}),
are represented as quadratic forms in the operators $B_{jk;s}$ equal to the identity matrix.
Consider the following trivial transformation: replace all inhomogeneous diagonal-element
constraints with a single inhomogeneous constraint given by the sum of all quadratic forms corresponding to the diagonal elements,
together with the remaining homogeneous constraints for all off-diagonal elements
and the differences between the two quadratic forms corresponding to the $j$ and $j-1$ diagonal elements.
This yields exactly the same set of constraints as before,
but with only a single inhomogeneous constraint;
this is referred to as the simplified (partial) constraint in Appendix \ref{RepresentationQC}.
Next, replace this single inhomogeneous constraint by the quadratic form in the denominator of
(\ref{fidelityProjectionsApproximationBBS}), set it equal to an arbitrary normalization constant:
\begin{align}
const&=\sum\limits_{s=0}^{N_s-1}\sum\limits_{j,j^{\prime}=0}^{D-1}\sum\limits_{k,k^{\prime}=0}^{n-1}
{B}^*_{jk;s}
Q_{jk;j^{\prime}k^{\prime}}
{B}_{j^{\prime}k^{\prime};s}
\label{NormDenom}
\end{align}
This transformation changes only the normalization of $B$,
since (\ref{fidelityProjectionsApproximationBBS}) is invariant under the scaling $B\to const \cdot B$.
However, with this new single inhomogeneous constraint,
the resulting SDP problem maximizes the objective function (\ref{fidelityProjectionsApproximationBBS}),
which can now be represented as a ratio of two quadratic forms.
Equation (\ref{NormDenom}) now replaces the simplified constraint (\ref{L2norm}).
The full set of constraints in (\ref{constraintKraussSpurBSKK}) is converted into (\ref{NormDenom})
together with the remaining $n(n+1)/2-1$ homogeneous constraints.
For (\ref{constraintKraussSpurBSJJ}), the constraints become (\ref{NormDenom})
together with the remaining $D(D+1)/2-1$ homogeneous constraints.
These constraints are then used to construct the $A_c$ matrices in the SDP problem (\ref{SDPconstraintsTr}).
The situation is analogous to Rayleigh quotient optimization, where the unconstrained optimization over $\psi$
\begin{align}
\frac{\Braket{\psi|S|\psi}}
{\Braket{\psi|Q|\psi}} &\to \max
\end{align}
is replaced by the optimization of a quadratic form
\begin{align}
\Braket{\psi|S|\psi} &\to \max
\end{align}
subject to a single quadratic constraint
\begin{align}
\Braket{\psi|Q|\psi} &=1
\end{align}
The Rayleigh quotient optimization problem can also be reduced to the SDP problem (\ref{SDPobjective}),
now using the density matrix instead of the Choi matrix.
However, for Rayleigh quotient optimization, a much more efficient method exists: the generalized eigenvalue problem
$\left|S|\psi\right>=\lambda\left|Q|\psi\right>$.

We approached this ``algebraic'' route in \cite{belov2025superstatePRE} by generalizing the eigenvalue problem to a new algebraic problem,
for $N_s=1$ it is:
\begin{align}
  S \mathcal{U} &= \lambda \mathcal{U}
  \label{eigenvaluesLikeProblem}
\end{align}
The superoperator $S$ is a tensor
$S_{jk;j^{\prime}k^{\prime}}$, 
the ``eigenvector'' $\mathcal{U}$ is a partially unitary operator
$\mathcal{U}_{jk}$,
and the ``eigenvalue'' $\lambda$ is a Hermitian $D \times D$ matrix of Lagrange multipliers  $\lambda_{ij}$,
an ``eigenmatrix'' instead of the usual eigenvalue-scalar.

We conducted numerical testing for projective operator reconstruction in a manner similar to unitary learning in Section \ref{UnitaryMappingReconstruction}. A projective operator $P$
was created by taking the first  $D$
rows from a unitary operator generated from random data.
Then, for randomly generated $\psi^{(l)}$, we applied the operator $P$
and normalized the result to the state 
$\phi^{(l)}$ (\ref{mappingProjPsiPhi}).
This follows a typical form of the data in ML and AI: normalized vectors of different dimensions.
Next, the tensors
$S_{jk;j^{\prime}k^{\prime}}$ and $Q_{jk;j^{\prime}k^{\prime}}$ 
were created, and the corresponding optimization problem with the fidelity in (\ref{fidelityProjectionsApproximationBBS})
was solved. In all cases, the solution was the Choi matrix of rank one,
and we selected the corresponding eigenvector.
It matched exactly the projective operator $P$ that was used to generate the sample.
Both \texttt{CVXPY} and \texttt{SDPA} were able to find the solution,
and it always matched the original operator $P$.
If we were to plot the fidelity and rank for a fixed $n$ and varied $1\le D\le n$,
it would resemble the exact reconstruction for unitary learning shown in Fig. \ref{resRanknUnitaryMapping}:
rank one and fidelity equal to one. This contrasts with the difficulties discussed in \cite{belov2024partiallyPRE},
Section V, where using a fidelity of a form other than (\ref{fidelityProjectionsApproximationBBS})
produced artifacts in the reconstruction for  $D<n$,
while no artifacts were observed for unitary learning with $D=n$.
See the class \texttt{\seqsplit{com/polytechnik/sdpexternalsolvers/TestChoiMatrixProjections.java}},
which demonstrates the reconstruction of projective operator 
from the sample in (\ref{mappingProjPsiPhi})
using the fidelity (\ref{fidelityProjectionsApproximationBBS}).

In this section, we formulate an optimization problem that allows for the reconstruction of projective operators.
The main difficulty in solving this problem lies in the choice of fidelity.
In Section V of \cite{belov2024partiallyPRE}, we demonstrated that for polynomial mapping data
(which is then transformed into quantum-channel input/output data using (\ref{psiXFGflocalizedAppendix})),
for a pure-state-type fidelity of the form $|\Braket{\phi|P|\psi}|^2$,
the map is reconstructed exactly when the input and output dimensions are equal.
When $D<n$, the obtained operator does not reconstruct the original projective operator $P$;
instead, the fidelity global maximum is reached for a projection different from the original $P$.
In Table I of \cite{belov2024quantumPRE}, a similar fidelity-choice problem is demonstrated for quantum channels,
where we formulated the major requirement for the fidelity: the extremum should be reached at the original mapping.
This requirement is analogous to Uhlmann's theorem for quantum-channel fidelity, but applied to maps that are CP and no longer TP.
We do not currently have a formal mathematical proof that the fidelity in the form (\ref{fidelityProjectionsApproximationBBS}) satisfies this criterion for the reconstruction of projective operators.
However, we do have global optimization algorithms -- either the iterative approach of Ref. \cite{belov2024partiallyPRE}
or the SDP convex optimization method developed in this paper -- that allow this condition to be tested numerically.
In thousands of trials, we randomly generated projective operators and applied them to random input data
to obtain the corresponding outputs. We then used this input/output data to reconstruct the projective operators.
As discussed above, in all tested cases the projective operators were reconstructed exactly.
Exact reconstruction (in all tests we performed) was achieved only for the fidelity (\ref{fidelityProjectionsApproximationBBS}).
Other quadratic forms of fidelity that we tested failed to satisfy this condition.
This leads us to conclude that the proposed form (\ref{fidelityProjectionsApproximationBBS}),
which is a ratio of two quadratic forms in the mapping operators,
allows for the exact reconstruction of projective operators.
A formal mathematical proof of this numerically established result remains a subject for future research.

A natural extension of projective operators is a higher-rank Kraus CP quantum channel,
which converts a unit matrix to a unit matrix, i.e., the constraints (\ref{constraintKraussSpurBSJJ}) with $N_s>1$.
In ML/AI, the IN/OUT data is typically normalized to 1, similar to (\ref{mappingProjPsiPhi}) for projective operators.
In this case, the fidelity in (\ref{fidelityProjectionsApproximationBBS}) allows for the reconstruction
of ``projective-type'' quantum channels. This form of knowledge representation can be highly advantageous for ML.

In this section, we demonstrate the application of the theory to reconstruct projective operators from data.
This is possible if we approximate the fidelity as a ratio of two quadratic forms in the mapping operators,
as shown in (\ref{fidelityProjectionsApproximationBBS}).
This is likely the most general form of the fidelity to which our theory can be applied.
This represents an important generalization of the quadratic fidelity in mapping operators (\ref{FidelityBS})
that we previously considered.
The fidelity form in (\ref{fidelityProjectionsApproximationBBS})
enables the construction of a much broader variety of proxy fidelities (approximations).
In our previous work \cite{belov2024quantumPRE}, when the proper fidelity in the form of (\ref{FidelityBS})
could not be constructed, we used approximations in the same form (\ref{FidelityBS})
and highlighted the limitations of such approximations.
The newly suggested form in (\ref{fidelityProjectionsApproximationBBS})
significantly expands the range of possible approximations that can be used,
facilitating the reconstruction of quantum channels of various forms from a broader set of sample data.

\section{\label{conclusion}Conclusion}

In this work we demonstrated that,
when the fidelity is a ratio of two quadratic functions in mapping operators,
a powerful SDP technique can be applied to reconstruct the quantum channel from data.
Commercial SDP optimization software is directly applicable to problems with Choi matrix dimensions under $1000 \times 1000$.
The optimization problem typically results in a low-rank Choi matrix,
typically under a few percent of the matrix dimension.
This low Kraus rank property has deep significance in AI and ML applications:
a relatively small Kraus rank quantum channel is typically sufficient to describe experimentally observed data.

The paper considers the problem of SDP-based quantum-channel reconstruction from both theoretical
and practical application viewpoints.
The two practical problems to which the theory can be readily applied are:
1). the approximate reconstruction of a CPTP quantum channel from a pure-state-to-pure-state map (\ref{psiXFGflocalizedAppendix}),
this type of map is typical in ML/AI
2). the reconstruction of a projective operator (Section \ref{projective}), which is not a CPTP map but represents an important practical problem; the exact reconstruction is possible with SDP when the fidelity is taken in the form of a ratio of two quadratic forms (\ref{fidelityProjectionsApproximationBBS}).

A quantum channel learning representation can be considered
a natural extension of unitary learning\cite{bisio2010optimal,arjovsky2016unitary,hyland2017learning,huang2021learning,yu2023optimal,belov2024partiallyPRE,belov2024quantumPRE};
the convex nature of the optimization problem makes this approach especially appealing for practical applications.

This work considers the reconstruction of a single quantum channel from classical mapping data.
The mapping itself can be seen as a form of computation.
A quantum computation algorithm consists of a sequence of unitary transformations.
For a given initial state, the quantum system's unitary evolution produces the desired computational result.
In ML/AI, unitary learning is typically used differently:
the input state is transformed into an output, and the specific components of the output state are considered
the features of interest that the ML system predicts.
Contrary to quantum computations, where simple gates are used
(as it is assumed that simpler gates are easier to implement in a physical system),
these classical unitary mappings are not necessarily required to be simple.
The quantum channel transformation (\ref{KrausOperator}) can be considered a form of computation, one way or another.

In this work, the reconstruction of a single quantum channel (\ref{qcClassicQ}) from IN/OUT classical data has been considered.
In \cite{belov2025superstatePRE}, we introduced the hierarchical concept of a density matrix network,
where a large quantum channel is represented as hierarchical transformations by simpler quantum channels.
Similar to how perceptron weights, combined with an activation function,
create specialized components that are then combined into a neural network,
we can consider a hierarchical quantum channel in a similar way.
Whereas for neural networks the choice of an appropriate topology is an important part of the model itself,
for a quantum channel represented as a hierarchy of simpler quantum-channel transformations,
the chosen topology (i.e., the form of the hierarchy) serves primarily as
a means of optimization rather than as a determinant of the represented mapping.
An additional advantage is that, while the internal behavior of neural networks is often poorly understood --
especially with respect to gauge invariance, i.e., which transformations of the input attributes leave the output unchanged --
the corresponding class of transformations that do not affect the result is known for quantum channels.

Importantly, since the SDP problem is convex,
the optimization of the total quantum channel can be performed as an iteration over individual, much smaller components.
This ``incremental learning'' feature is seldom in neural networks,
it is typically available only for specific topologies.
Since the SDP problem for quantum channels is convex,
this incremental learning feature is available for an arbitrary quantum channel.
The specific topology chosen for the density matrix network is simply a matter of computational optimization.
For smaller dimensions, the problem can always be optimized directly without assuming any particular topology,
using the SDP technique we considered in this paper.
This contrasts with neural networks, where selecting the proper topology is a crucial factor for success.
We believe that representing knowledge in the form of a quantum channel allows for modeling a wide variety of data types.
The observed low rank of the reconstructed Choi matrix leads us to conclude that low Kraus rank quantum channels are typically sufficient to map classical data.
It is important to note that the developed approach can also be applied to other forms of quantum channels,
such as projective operators.

\begin{acknowledgments}
This research was supported by Autretech Group,
a resident company of the Skolkovo Technopark.
  We thank our colleagues from the Autretech R\&D department
  who provided insight and expertise that greatly assisted the research.
  Our grateful thanks are also extended
  to Mr. Gennady Belov for his methodological support in doing the data analysis.

Vladislav Malyshkin is grateful to Yaroslav Mikhailovich Beltukov for stimulating discussions
on the applications of Semidefinite Programming to physical problems,
and to Eugenius Levovich Ivchenko for his comments regarding the requirement
of a simple demonstration of the technique,
which is now addressed in Section \ref{UnitaryMappingReconstruction}.

The article was written based on the presentation at the
colloquium\cite{iipColloq2025}
on the occasion of the 96th anniversary
of Iya Pavlovna Ipatova's\cite{birman2004ija} birthday, held on December 19, 2025.
\end{acknowledgments}

\appendix

\section{\label{RepresentationQC}Quantum Channel Representation in Lower Diagonal Matrix Form}

The optimization problem we consider is to maximize the total fidelity (\ref{FidelityBS}) over the space of Kraus operators
${B}_{s}$ (\ref{KrausOperator}), subject to the constraints (\ref{constraintKraussSpurBSKK}) or (\ref{constraintKraussSpurBSJJ}).
The main idea behind our algebraic technique for numerical computations
is to replace the quadratic constraints (\ref{constraintKraussSpurBSKK}) or (\ref{constraintKraussSpurBSJJ})
with a single simplified (partial) quadratic constraint, which is typically the sum of the diagonal elements,
and a set of convergence-helper homogeneous linear constraints derived from the remaining quadratic constraints.
This allows us to employ the eigenvalue problem as the main iterative step,
with the convergence-helper constraints reducing the size of the vector space.
For unitary mapping, vectorizing the unitary operator provides a proper parametrization of the search space.
The problem is non-convex, but multiple solution candidates (eigenvectors)
greatly increase the chances of finding the global maximum compared to
methods based on gradient- or Newton-type algorithms,
which generate only a single solution candidate.

For a quantum channel (\ref{KrausOperator}), a similar approach was developed in \cite{belov2024quantumPRE}.
There, we considered the quantum channel in its canonical representation,
$\mathrm{Tr}B_s B^{\dagger}_{s^{\prime}}\sim\delta_{ss^{\prime}}$.
This required an additional $N_s(N_s-1)/2$ constraints and corresponding Lagrange multipliers.
The updated version of the algorithm presented below uses the Cholesky decomposition to eliminate these extra constraints.
Now, a requirement for some components of the $B_s$ matrix to be zero arises explicitly,
which allows for the elimination of the extra constraints,
thus avoiding an increase in the search space and the need for additional Lagrange multipliers.
The idea is to parametrize $\mathcal{J}$ (\ref{Jmatrix}) as a product of two matrices
\begin{align}
\mathcal{J}&=B^* B^{T} \label{LowerDiagParametrization}
\end{align}
where $B_{jk;s}$ is now a lower-diagonal matrix of dimension $Dn\times N_s$.
This form is actually Eq. (\ref{Jmatrix}), with the ``gauge'' for $B_{jk;s}$ being a lower-diagonal matrix.
By choosing $N_s$,
we can construct a quantum channel of any required Kraus rank $N_s$.
The matrix elements $B_{jk;s}$ are nonzero only for $s\le j n+k$ with $s<N_s$.
From a computational perspective, an important property is that the full trace
$\mathrm{Tr} B^* B^{T}$ reduces to a simple Frobenius norm:
\begin{align}
|B|^2&=
\sum\limits_{s=0}^{N_s-1}
\sum\limits_{j=0}^{D-1}
\sum\limits_{k=0}^{n-1} |B_{jk;s}|^2
\label{L2norm}
\end{align}
This gives a vector-style $\mathrm{L}^2$ norm for the operator $B$.
When all elements of $B$  are stored consecutively in a single vector (row by row),
the full-trace operation becomes ordinary vector scalar products.
\begin{align}
\Braket{B|\breve{B}}&= \mathrm{Tr} B^{*} \breve{B}^{T}=
\sum\limits_{s=0}^{N_s-1}
\sum\limits_{j=0}^{D-1}
\sum\limits_{k=0}^{n-1} B_{jk;s}^{*} \breve{B}_{jk;s}
\label{Lspace}
\end{align}
This $LL^T$ (Cholesky-type) decomposition is often used to parametrize positive semidefinite matrices.
Whereas the problem in the form (\ref{SDPobjective}) subject to (\ref{SDPconstraintsTr})
and CP constraints (\ref{SDPconstraint}) is a convex SDP optimization problem \cite{boyd2004convex},
the $LL^T$ transformation automatically satisfies only the CP constraints.
The actual problem written in this new basis may not always retain the convexity property.
In some cases, a convexity-like behavior may still be observed,
and an arbitrary optimization method can effectively find the solution to the convex optimization problem.
In our case, we used eigenvalue-based optimization, which is not particularly sensitive to problem convexity.
For any individual problem, the convexity property in the $LL^T$ basis should be initially checked to decide on the appropriate optimization method to apply.

In the form (\ref{LowerDiagParametrization}) with $N_s\le Dn$,
this representation allows us to efficiently parametrize $\mathcal{J}$ for any desired Kraus rank $N_s$.
Unitary learning corresponds to $N_s=1$, with the sought operator  $\mathcal{U}$ stored in a single column.
A quantum channel of Kraus rank three corresponds to a lower diagonal $B$ matrix with $Dn$ rows and $N_s=3$ columns of lengths
$Dn$, $Dn-1$, and $Dn-2$, respectively.
\begin{align}
B&=
\underset
{
 \underbrace{\hphantom{\begin{matrix}\star & \star & \star \end{matrix}}}_{N_s}
 \hphantom{0}
 \underbrace{\hphantom{\begin{matrix}0 & \dots &0\end{matrix}}}_{Dn-N_s}
}
{
\left(
  \begin{matrix}
  \star & 0 & 0  & 0 & \dots &0 \\
  \star & \star & 0 &0 & \dots &0 \\
  \star & \star & \star &0 & \dots &0  \\
  \star & \star & \star &0 & \dots &0  \\
  \vdots  & \vdots  & \vdots &\vdots &  & \vdots \\
  \star & \star & \star &0  &\dots &0 
  \end{matrix}
\right)
}
\label{matrixView3}
\end{align}
From a computational viewpoint, an eigenproblem (obtained using the simplified, partial constraint)
in the space of operators $B$
becomes an ordinary eigenproblem for a vector of dimension $DnN_s - N_s(N_s-1)/2$, i.e.,
\begin{align}
\dim&=\frac{(2Dn-N_s+1)N_s}{2} \label{dimProblem}
\end{align}
which is the number of non-zero elements in (\ref{matrixView3}).
The major difference from the SDP problem discussed above -- which always assumes $N_s=Dn$ --
is that here we can construct a solution with a pre-specified Kraus rank $N_s$.
Imposing a specific rank on $\mathcal{J}$ in addition to the PSD condition $\mathcal{J} \succeq 0$ (\ref{SDPconstraint})
makes the problem non-convex and significantly more difficult.

Importantly, from the perspective of numerical implementation,
that the representation (\ref{LowerDiagParametrization})
creates a problem very similar to the one we previously considered.
The only difference from the algorithm in \cite{belov2024partiallyPRE} lies in the increased dimension of the eigenproblem,
which becomes (\ref{dimKK}) for constraint (\ref{constraintKraussSpurBSKK}) and (\ref{dimJJ}) for constraint (\ref{constraintKraussSpurBSJJ}) when the convergence-helper linear constraints are taken into account.
\begin{align}
\dim_{\mathrm{Tr}}&=\frac{(2Dn-N_s+1)N_s}{2} - \frac{n(n+1)}{2} +1 \label{dimKK} \\
\dim_{\mathds{1}}&=\frac{(2Dn-N_s+1)N_s}{2} - \frac{D(D+1)}{2} +1 \label{dimJJ}
\end{align}
This illustrates the main strengths and weaknesses of the approach.
Whereas for unitary learning ($D=n$, $N_s=1$) the eigenproblem dimension grows as $O(n^2)$,
for a full-rank quantum channel ($N_s=Dn$) the eigenproblem dimension grows as $O(n^4)$
when $D=n$.
However, as discussed in Section \ref{Reconst}, the latter problem is convex and can be solved using other optimization methods.
The case of a low-Kraus-rank quantum channel (low $N_s$)
is more difficult and represents the main application problem for the algorithm we developed in \cite{belov2024partiallyPRE}.

The optimization problem consists in Lagrangian optimization.
In this appendix, we consider the unit input matrix to unit output matrix constraint (\ref{constraintKraussSpurBSJJ}),
\begin{align}
G_{jj^{\prime}}(B,\breve{B)}&=\sum\limits_{s=0}^{N_s-1} \sum\limits_{k=0}^{n-1} B_{jk;s}^{*}\breve{B}_{j^{\prime}k;s} \label{GramGJJ} \\
\delta_{jj^{\prime}}&=G_{jj^{\prime}}(B,B)
\label{constraintKraussSpurLSJJ}
\end{align}
for the solution $B_{jk;s}$ (\ref{matrixView3})
as it is somewhat easier to handle numerically.
The constraints are analogous to (\ref{Lspace}), but now the partial trace is used in (\ref{GramGJJ}).
The formulas corresponding to the trace preservation constraints
(\ref{constraintKraussSpurBSKK}) are presented in Appendix \ref{ConstraintsTracePreserving} below.
Introduce a Lagrangian.
\begin{align}
  \mathcal{L}&=
  \sum\limits_{s=0}^{N_s-1}
  \sum\limits_{j,j^{\prime}=0}^{D-1}\sum\limits_{k,k^{\prime}=0}^{n-1}
             B^*_{jk;s}S_{jk;j^{\prime}k^{\prime}}B_{j^{\prime}k^{\prime};s} \label{KrausLagrangianJJ} \\
             &+
   \sum\limits_{j,j^{\prime}=0}^{D-1}        
   \lambda_{jj^{\prime}}\left[\delta_{jj^{\prime}}-
\sum\limits_{s=0}^{N_s-1}
\sum\limits_{k^{\prime}=0}^{n-1}B^*_{jk^{\prime};s} B_{j^{\prime}k^{\prime};s} \right] 
   \nonumber
\end{align}
Note that in (\ref{FidelityBS}), the tensor $S_{jk;j^{\prime}k^{\prime}}$ appears exactly the same
for all $s$ and for all Kraus operators $B_s$.
This leads to degeneracy in the optimization problem.
However, if instead of general $B$ we use lower lower diagonal $B_{jk;s}$ from (\ref{matrixView3}),
this degeneracy does not occur, because for each fixed $s$ the corresponding vector $B_{jk;s}$
contains a different number of non-zero elements
(indexed by the scalar index $jn+k$, corresponding to a row in (\ref{matrixView3}) and column $s$).
While the first term in (\ref{KrausLagrangianJJ}) may appear similar to (\ref{FidelityBS}), it is in fact different, because it eliminates the upper-diagonal terms in $B_{jk;s}$ (\ref{matrixView3}).
In (\ref{KrausLagrangianJJ}), one may think of $S_{jk;j^{\prime}k^{\prime}}$
as effectively depending on $s$ through the condition $s\le\min(jn+k,N_s-1)$, i.e.,
as $S_{s,jk;s^{\prime},j^{\prime}k^{\prime}}$, which is necessary to avoid the degeneracy of the problem discussed in \cite{belov2024quantumPRE}.
This is an important feature of the Cholesky decomposition (\ref{LowerDiagParametrization}).

In the $N_s=1$ case, varying (\ref{KrausLagrangianJJ}) yields the familiar Lagrange stationarity condition, which results in the algebraic problem introduced in \cite{belov2024partiallyPRE}:
\begin{align}
S \mathcal{U}&=\lambda \mathcal{U} \label{UregJJ}
\end{align}
In the $N_s>1$ case a similar algebraic problem with respect to lower diagonal $B$ can be obtained, but it's analysis requires a separate study.

An important idea of the numerical method is the introduction of a simplified (partial) constraint,
which is the sum of all diagonal elements of (\ref{GramGJJ}); this yields (\ref{L2norm}) with $|B|^2 = D$.
To obtain the optimization problem
when using simplified (partial) constraints instead of the proper ones,
the variation of $\mathcal{L}$ gives the eigenvectors of the matrix $\mathcal{S}_{jk;j^{\prime}k^{\prime}}$
\begin{align}
  &\mathcal{S}_{jk;j^{\prime}k^{\prime}}=
             S_{jk;j^{\prime}k^{\prime}}
             -\lambda_{jj^{\prime}}\delta_{kk^{\prime}} \label{SJJ}\\
             &
\frac{
\sum\limits_{s=0}^{N_s-1}    
\sum\limits_{j,j^{\prime}=0}^{D-1}\sum\limits_{k,k^{\prime}=0}^{n-1}
             B^*_{jk;s}\mathcal{S}_{jk;j^{\prime}k^{\prime}}B_{j^{\prime}k^{\prime};s}
             }
             {
\sum\limits_{s=0}^{N_s-1}  
\sum\limits_{j=0}^{D-1}
\sum\limits_{k=0}^{n-1}\left|B_{jk;s}\right|^2
             }
   \xrightarrow[B]{\quad }\max
   \label{KrausOptimizationVarSJJ} \\
&
\sum\limits_{j,j^{\prime}=0}^{D-1}\sum\limits_{k,k^{\prime}=0}^{n-1}
   \mathcal{S}_{jk;j^{\prime}k^{\prime}}B^{[i]}_{j^{\prime}k^{\prime};s}
   =\mu^{[i]} B^{[i]}_{jk;s}
   \label{MuEigenproblem}
\end{align}
The problem (\ref{MuEigenproblem}) is an eigenproblem of dimension (\ref{dimProblem}),
since only the non-zero terms with $s\le \min(jn+k,N_s-1)$ are retained.
For numerical implementation, one can think of $B_{jk;s}$ as a vector $B_i$ of length (\ref{dimProblem}),
where the scalarized multi-index $i$ ranges from $0$ to $\mathrm{dim}-1$.

As discussed in Section \ref{projective}, for reconstructing projective operators,
the fidelity in (\ref{fidelityProjectionsApproximationBBS}) can be expressed in the form where,
instead of using the simplified constraint (\ref{L2norm}), one may use the denominator in (\ref{fidelityProjectionsApproximationBBS}).
This leads to a slight modification in the formulas.
This form results in a generalized eigenproblem similar to (\ref{MuEigenproblem}).
The formula for the Lagrange multipliers $\lambda_{jj^{\prime}}$, as given in (\ref{lambdaFromU}),
is also slightly modified to include $Q_{jk;j^{\prime}k^{\prime}}$.

As discussed in \cite{belov2024partiallyPRE,belov2024quantumPRE},
the direct optimization of (\ref{KrausOptimizationVarSJJ}), which leads to an eigenproblem of dimension (\ref{dimProblem}),
does not guarantee algorithm convergence.
To reduce the basis dimension to (\ref{dimJJ}) and achieve convergence,
it is necessary to include convergence-helper homogeneous linear constraints
derived from the quadratic constraints (\ref{constraintKraussSpurLSJJ}).
See Appendix A.5 of \cite{belov2024quantumPRE} for the formulas and
\texttt{\seqsplit{com/polytechnik/kgo/LLTbasis.java:getHelperConstraintsOffdiag0DiagEq}}
for an updated numerical implementation.
The idea is simple: for a given solution candidate $B$, find $D(D+1)/2-1$ constraints $C^{(m)}$ such that,
for an arbitrary $\breve{B}$, $D(D-1)/2$ of them will have $\Braket{C^{(m)}|\breve{B}}$ from (\ref{Lspace})
equal to the sum of two off-diagonal ($j_m\ne j^{\prime}_m$)  elements $G_{j_m\,j^{\prime}_m}(B,\breve{B})+G^*_{j^{\prime}_m\,j_m}(B,\breve{B})$ from (\ref{GramGJJ}),
and $D-1$ of them will have $\Braket{C^{(m)}|\breve{B}}$
equal to the difference of two diagonal elements: 
$G_{j_m\,j_m}(B,\breve{B})-G_{j_m-1\,j_m-1}(B,\breve{B})$.

After this, at the eigenvalue-selection step of the reduced-size (\ref{dimJJ}) eigenproblem, one needs to choose the
eigenvector $B^{[i]}_{jk;s}$ corresponding to the maximal eigenvalue.
Numerical experiments show that this choice leads to convergence to the global maximum,
especially in the unitary case.

The obtained solution $B_{jk;s}$  may not satisfy the full set of constraints (\ref{constraintKraussSpurLSJJ});
only the partial constraint (\ref{L2norm}) is always satisfied.
To adjust the solution, we apply the $G^{-1/2}$ transform (see Ref. \cite{belov2024quantumPRE}, Eq. (A23)),
which then yields the fully constrained solution:
\begin{align}
\widetilde{B}_{jk;s}&=
\sum\limits_{j^{\prime}=0}^{D-1}G^{-1/2}_{jj^{\prime}}B_{j^{\prime}k;s}
\label{G05transformJJ}
\end{align}
Here $G^{-1/2}_{jj^{\prime}}$ denotes the inverse square root of the Gram matrix (\ref{GramGJJ}),
calculated at the current iteration $B_{jk;s}$.
If the Gram matrix has a zero eigenvalue, the adjustment procedure will fail.
For some combinations of $N_s,D,n$, the constraints cannot be satisfied for an arbitrary $B$.
For example, when $N_s=1$ and $D<n$ in the case of the trace-preserving constraints (\ref{constraintKraussSpurBSKK}),
the matrix $G$ (\ref{GramGKK}) has at least $n-D$ zero eigenvalues.
We should also note that applying $G^{-1/2}_{jj^{\prime}}$ (\ref{G05transformJJ}) could potentially cause $\widetilde{B}_{jk;s}$
to no longer belong to the set of lower diagonal matrices (\ref{matrixView3}).
This issue is technical and can be readily resolved.
The proposed transform allows for the conversion of an arbitrary quantum channel $B_{jk;s}$
(\ref{KrausOperator}) to an entirely equivalent form of lower-diagonal matrices (\ref{matrixView3}).
For a similar transformation to canonical form, see Appendix A.1 of Ref. \cite{belov2024quantumPRE}.
Note that in (\ref{LowerDiagParametrization}), one can write it as
\begin{align}
\mathcal{J}&=B^* Q Q^{\dagger} B^{T} \label{LowerDiagParametrizationQQ}
\end{align}
where $Q$ is an arbitrary unitary operator of dimension $N_s$.
Define an $N_s\times N_s$ matrix $M_{r;s}$
by setting its elements equal to $\widetilde{B}_{jk;s}$
corresponding to the upper-left $N_s\times N_s$ block in (\ref{matrixView3});
for $\widetilde{B}_{jk;s}$ from (\ref{G05transformJJ}) the matrix $M$ is not necessarily lower diagonal.
Applying the \href{https://en.wikipedia.org/wiki/QR_decomposition}{QR decomposition} to this matrix, we obtain (for the real space):
\begin{align}
M^T&=Q R \label{QRdecomposition}
\end{align}
where $Q$ is the orthogonal matrix and $R$ is the upper diagonal matrix.
From here, it follows that the matrix $Q$ will transform $\widetilde{B}_{jk;s}$
into the lower diagonal form of (\ref{matrixView3}).
See the class
\texttt{\seqsplit{com/polytechnik/kgo/LLTbasis.java:ConvertStateToLowerDiagMatrix}},
which uses QR decomposition to obtain $B$ in the lower diagonal form that can be used as a vector.
This transform is an identity and does not change $\mathcal{J}$ (\ref{LowerDiagParametrizationQQ}).
In terms of Kraus-operator transformations, this is a gauge transformation that mixes the $N_s$
Kraus operators without changing the underlying quantum channel (\ref{KrausOperator}).
\begin{align}
B_{r}=\sum\limits_{s=0}^{N_s-1}Q_{rs} \widetilde{B}_{s}
\label{BsIdentity}
\end{align}

The final step of the iteration is to update the Lagrange multipliers matrix $\lambda_{ji}$.
Following \cite{belov2025superstatePRE}, we obtain the expression:
\begin{align}
\lambda_{ji}&=\mathrm{Herm}\sum\limits_{s=0}^{N_s-1}\sum\limits_{j^{\prime}=0}^{D-1}\sum\limits_{k,k^{\prime}=0}^{n-1}B^{*}_{ik;s}S_{jk;j^{\prime}k^{\prime}}B_{j^{\prime}k^{\prime};s}
\label{lambdaFromU}
\end{align}
which can also be obtained directly by multiplying $\frac{\delta\mathcal{L}}{\delta B^{*}_{jk;s}}$
by $B^{*}_{ik;s}$
and summing over $s$ and $k$, without considering a dual problem.
The expression (\ref{lambdaFromU}) generalizes the standard eigenvalue relation $\lambda = \Braket{\psi|H|\psi}$.

The current version of the algorithm differs from the algorithm in Ref. \cite{belov2024quantumPRE}
in only one significant respect: it now uses the Cholesky decomposition (\ref{LowerDiagParametrization}) to represent the state.
This allows us, in the $N_s>1$ case,
to eliminate problem degeneracy and to require only a \textsl{single} solution adjustment (\ref{G05transformJJ}).
The updated version uses $N_s(N_s-1)/2$ fewer variables than the version described in \cite{belov2024quantumPRE};
therefore, no ``second set'' of Lagrange multipliers $\nu_{ss^{\prime}}$ is required,
as in Eq. (A2) of Ref. \cite{belov2024quantumPRE}.
As long as we seek only the global solution of the optimization problem (\ref{KrausLagrangianJJ})
(that is, no ``external'' linear constraints are used, which would be required to build a hierarchy of solutions),
only a single solution adjustment (\ref{G05transformJJ}) is applied, thereby avoiding two potentially conflicting transformations.
In the $N_s=1$ case, the described algorithm is identical to that in \cite{belov2024partiallyPRE}.

\subsection{\label{ConstraintsTracePreserving}Trace-Preserving Constraints: Mathematical Formulations and Equations}

The most commonly considered constraint for a quantum channel map (\ref{KrausOperator}) is trace preservation (\ref{constraintKraussSpurBSKK}).
This is an alternative to (\ref{constraintKraussSpurBSJJ}), which enforces the constraint that the unit matrix
$A^{IN}$
is mapped to the unit matrix $A^{OUT}$.
The technical meaning of this constraint is to construct a unit matrix of dimension $n\times n$
from a sum of $N_s$ matrices of the same dimension, each with rank $D$ (or lower).
For a unitary mapping with $N_s=1$ and $D=n$, both forms of constraints coincide.
For the trace-preservation constraints (\ref{constraintKraussSpurBSKK}), the following modifications apply.

Instead of (\ref{KrausLagrangianJJ}),
the Lagrangian corresponding to the trace preservation constraints (\ref{constraintKraussSpurBSKKChoi}) is
\begin{align}
  \mathcal{L}&=
  \sum\limits_{s=0}^{N_s-1}
  \sum\limits_{j,j^{\prime}=0}^{D-1}\sum\limits_{k,k^{\prime}=0}^{n-1}
             B^*_{jk;s}S_{jk;j^{\prime}k^{\prime}}B_{j^{\prime}k^{\prime};s} \label{KrausLagrangianKK} \\
             &+
\sum\limits_{k,k^{\prime}=0}^{n-1}        
   \lambda_{kk^{\prime}}\left[\delta_{kk^{\prime}}-
\sum\limits_{s=0}^{N_s-1}
\sum\limits_{j^{\prime}=0}^{D-1}B^*_{j^{\prime}k;s} B_{j^{\prime}k^{\prime};s} \right] 
   \nonumber
\end{align}
In the $N_s=1$ case, varying (\ref{KrausLagrangianKK}) yields the familiar Lagrange stationarity condition,
which results in the algebraic problem:
\begin{align}
S\mathcal{U}&=\mathcal{U}\lambda  \label{UregKK}
\end{align}
Instead of (\ref{SJJ}), the variation of $\mathcal{L}$ (\ref{KrausLagrangianKK})
yields a matrix from which the solution candidates can be obtained as its eigenvectors:
\begin{align}
  \mathcal{S}_{jk;j^{\prime}k^{\prime}}=&
             S_{jk;j^{\prime}k^{\prime}}
             -\lambda_{kk^{\prime}}\delta_{jj^{\prime}} \label{SKK}
\end{align}
Instead of (\ref{G05transformJJ}) the solution adjustment transform now is:
\begin{align}
G_{kk^{\prime}}&=\sum\limits_{s=0}^{N_s-1} \sum\limits_{j=0}^{D-1} B_{jk;s}^{*}B_{jk^{\prime};s} \label{GramGKK}  \\
\widetilde{B}_{jk;s}&=
\sum\limits_{k^{\prime}=0}^{n-1}G^{-1/2}_{kk^{\prime}}B_{jk^{\prime};s}
\label{G05transformKK}
\end{align}
When the Kraus rank $N_s$ is insufficient, the Gram matrix (\ref{GramGKK}) may have zero eigenvalues,
and the transform (\ref{G05transformKK}) cannot be applied.
In practice, the matrix (\ref{GramGKK}) is more likely to have a zero eigenvalue than the matrix (\ref{GramGJJ})
that we considered above.
Also, similarly to the above, the obtained $\widetilde{B}_{jk;s}$
may not satisfy the lower diagonal condition (\ref{matrixView3}).
This issue can be resolved in the same way using the QR transform (\ref{QRdecomposition}).

Instead of (\ref{lambdaFromU}), the expression for the Lagrange multipliers for the next iteration is:
\begin{align}
\lambda_{kq}&=\mathrm{Herm}\sum\limits_{s=0}^{N_s-1}\sum\limits_{j,j^{\prime}=0}^{D-1}\sum\limits_{k^{\prime}=0}^{n-1}
B^*_{jq;s}S_{jk;j^{\prime}k^{\prime}}B_{j^{\prime}k^{\prime};s}
\label{lambdaFromUKK}
\end{align}
Everything else regarding the trace-preservation constraints (\ref{constraintKraussSpurBSKK})
is almost identical to the expressions presented above in Appendix \ref{RepresentationQC}
for the constraints (\ref{constraintKraussSpurBSJJ}).

\subsection{\label{NumImplementationNotes}Numerical Implementation Approach and Challenges}

For the updated version of this algorithm, see the class
\texttt{\seqsplit{com/polytechnik/kgo/QCInverseProblem.java}},
which implements the algorithm of above.
Compared to the older version,
\texttt{\seqsplit{com/polytechnik/kgo/KGOIterationalSubspaceLinearConstraintsB.java}},
it removes all the ``external'' constraints code (used to build a hierarchy of solutions)
and explicitly takes the value of the Kraus rank $N_s$ as an argument.
For $N_s=1$, it is almost identical to
\texttt{\seqsplit{com/polytechnik/kgo/KGOIterationalSubspaceLinearConstraintsB.java}} from \cite{belov2024quantumPRE} and
\texttt{\seqsplit{com/polytechnik/kgo/KGOIterationalSubspaceLinearConstraints.java}} from \cite{belov2024partiallyPRE},
with external constraints disabled.
Also, see the class \texttt{\seqsplit{com/polytechnik/kgo/LLTbasis.java}},
which now includes all operations in the  lower diagonal basis of the Cholesky decomposition (\ref{LowerDiagParametrization}).
The introduction of this class makes the implementation in
\texttt{\seqsplit{com/polytechnik/kgo/QCInverseProblem.java}} much cleaner.
Additionally, this class is used to create the matrices $A_c$ (\ref{SDPconstraintsTr}) when using commercial SDP software.

Whereas the physical applications are demonstrated and discussed in Section \ref{Demonstrations} above,
this appendix focuses only on the numerical aspects of the developed algorithm.
What allows the algorithm to find the global maximum is its use of the eigenproblem (\ref{KrausOptimizationVarSJJ})
as the core building block.
While derivative-based methods do not guarantee finding the global maximum,
the use of the eigenproblem guarantees finding the globally optimal solution in the space of dimension
$\dim_{\mathrm{Tr}}$ (\ref{dimKK}) for the constraints (\ref{constraintKraussSpurBSKK})
or $\dim_{\mathds{1}}$ (\ref{dimJJ}) for the constraints (\ref{constraintKraussSpurBSJJ}).
What prevents the algorithm from finding the global maximum in a single step and requires iterations
is the fact that the space in which we search for the solution is known only approximately.
Therefore, iterations are required.
This space is determined by the convergence-helper homogeneous linear constraints
derived from the quadratic constraints (\ref{constraintKraussSpurBSKK}) or (\ref{constraintKraussSpurBSJJ}).
For a real-valued space, there are $n(n+1)/2-1$ and $D(D+1)/2-1$ constraints, respectively.
Numerical experiments show the crucial importance of the convergence-helper constraints;
removing even a single one often prevents the algorithm from converging.

The numerical testing of the algorithm from Appendix \ref{RepresentationQC}
demonstrates its ability to reconstruct a quantum channel for relatively low values of $N_s$.
For $N_s \gtrsim 4$, the algorithm often iterates near Kraus rank-one solutions,
and it frequently takes a large number of iterations to escape from a Kraus rank-one local maximum.
The selection of the state corresponding to a non-maximal eigenvalue
at iteration step can improve the situation to some extent.

There is an important synergy between the described algorithm and SDP programming.
For the quantum channel reconstruction problem,
satisfaction of the constraints (\ref{constraintKraussSpurBSKKChoi}) or (\ref{constraintKraussSpurBSJJChoi})
can be readily obtained from the current iteration Choi matrix $\mathcal{J}_{jk;j^{\prime}k^{\prime}}$ (\ref{Jmatrix}).
First, the matrix  $G_{kk^{\prime}}$ (\ref{GramGKK}) or $G_{jj^{\prime}}$ (\ref{GramGJJ})
can be obtained as a partial trace of the current Choi matrix $\mathcal{J}_{jk;j^{\prime}k^{\prime}}$.
Then, by applying the transformation
\begin{align}
\widetilde{\mathcal{J}}_{jk;j^{\prime}k^{\prime}}&=\sum\limits_{k^{\prime\prime},k^{\prime\prime\prime}=0}^{n-1}
G^{-1/2}_{kk^{\prime\prime}}G^{-1/2}_{k^{\prime}k^{\prime\prime\prime}} \mathcal{J}_{jk^{\prime\prime};j^{\prime}k^{\prime\prime\prime}}
\label{ChoiTransformKK} \\
\widetilde{\mathcal{J}}_{jk;j^{\prime}k^{\prime}}&=\sum\limits_{j^{\prime\prime},j^{\prime\prime\prime}=0}^{D-1}
G^{-1/2}_{jj^{\prime\prime}}G^{-1/2}_{j^{\prime}j^{\prime\prime\prime}} \mathcal{J}_{j^{\prime\prime}k;j^{\prime\prime\prime}k^{\prime}}
\label{ChoiTransformJJ}
\end{align}
one can immediately obtain the Choi matrix satisfying the constraints
(\ref{constraintKraussSpurBSKKChoi}) or (\ref{constraintKraussSpurBSJJChoi}), respectively.
This transformation creates a ``minimal disturbance'' to the current iteration of $\mathcal{J}_{jk;j^{\prime}k^{\prime}}$
and allows for an efficient optimization of the SDP algorithm.

\section{\label{SoftwareDescription}Software Usage Description}

The software \cite{polynomialcode} is written in Java.
The related code is in the \texttt{\seqsplit{com/polytechnik/\{kgo,utils,sdpexternalsolvers\}}} directories.
To test the provided software, install Java 25 or later, install \texttt{CVXPY}\cite{diamond2016cvxpy} as \texttt{\seqsplit{pip3\ install\ cvxpy}},
and recompile and install \texttt{SDPA}\cite{yamashita2010high}  from \href{https://sourceforge.net/projects/sdpa/}{sdpa.sourceforge.net}.
Additionally install the Gson library from \href{https://github.com/google/gson}{github.com/google/gson}.

Download the source code \cite{polynomialcode} from the archive
\href{http://www.ioffe.ru/LNEPS/malyshkin/AMuseOfCashFlowAndLiquidityDeficit.zip}{\texttt{\seqsplit{AMuseOfCashFlowAndLiquidityDeficit.zip}}},
then decompress and recompile it.
Here, we use the backslash ``$\backslash$'' to split lines to fit the two-column PRE format;
BASH interprets it correctly, allowing the commands to be copied directly from the article into the BASH prompt.
\begin{verbatim}
unzip AMuseOfCashFlowAndLiquidityDeficit.zip
javac -g com/polytechnik/\
{kgo,utils,sdpexternalsolvers}/*java
\end{verbatim}
Make sure that the java Gson library, python \texttt{CVXPY} library\cite{diamond2016cvxpy},
and the JNI library \texttt{\seqsplit{com/polytechnik/sdpexternalsolvers/libSDPAsolver.so}}
linking to \texttt{SDPA}\cite{yamashita2010high} are installed/built.
The commands for building the JNI library and linking it to SDPA are provided
in the Javadoc section of \texttt{\seqsplit{com/polytechnik/sdpexternalsolvers/SDPAsolver.java}}.
After that one can run the provided software:
\begin{verbatim}
java -Djava.library.path=\
com/polytechnik/sdpexternalsolvers/ \
  com/polytechnik/sdpexternalsolvers/\
TestChoiMatrixUnitaryLearning
\end{verbatim}
See other programs in the classes \texttt{\seqsplit{com/polytechnik/sdpexternalsolvers/\{TestChoiMatrixProjections,TestChoiMatrixUnitaryLearning,TestChoiMatrixRankOnRandomSK,TestChoiMatrixRankOnSKFromRandomSample,TestChoiMatrixRankOnSKFromQuantumChannelOutPureAsMaxEVofRho\}.java}}.
Test usage is provided in the javadoc preamble.

The full codebase is available from \cite{polynomialcode},
file \href{http://www.ioffe.ru/LNEPS/malyshkin/AMuseOfCashFlowAndLiquidityDeficit.zip}{\texttt{\seqsplit{AMuseOfCashFlowAndLiquidityDeficit.zip}}}.
It is a large codebase that also includes other related projects and is expected to evolve
as the algorithms presented in this work evolve.
The supplemental materials \cite{EP12456supplemental} for this article contain the minimal code needed to reproduce
the results of this paper.
This code will not evolve; it corresponds exactly to the results presented in the published paper.

\bibliography{LD,mla}

@MISC{polynomialcode,
author={Malyshkin, Vladislav Gennadievich},
title={{The repository for ML and QML code}},
note={\url{http://www.ioffe.ru/LNEPS/malyshkin/code.html}.
The source code is available at
\href{http://www.ioffe.ru/LNEPS/malyshkin/AMuseOfCashFlowAndLiquidityDeficit.zip}{\texttt{\seqsplit{AMuseOfCashFlowAndLiquidityDeficit.zip}}}
},
year="2014-2026",
url={http://www.ioffe.ru/LNEPS/malyshkin/code.html}
}

@article{malyshkin2019radonnikodym,
  title={{On The Radon-Nikodym Spectral Approach With Optimal Clustering}},
  author={Malyshkin, Vladislav Gennadievich},
  journal={arXiv preprint arXiv:1906.00460},
  year={2019},
  url={https://arxiv.org/abs/1906.00460},
  doi={10.48550/arXiv.1906.00460}
}

@book{bernard2009moments,
  title={{Moments, positive polynomials and their applications}},
  author={Lasserre, Jean-Bernard},
  volume={1},
  year={2009},
  publisher={World Scientific},
  url={https://www.worldscientific.com/worldscibooks/10.1142/p665},
  doi={10.1142/p665}
}

@book{kraus1983states,
  title={{States, Effects, and Operations: Fundamental Notions of Quantum Theory}},
  author={Kraus, Karl},
  publisher={Springer-Verlag},
  volume={190},
  series={Lecture Notes in Physics},
  year={1983},
  note={{Lectures in Mathematical Physics at the University of Texas at Austin}},
  isbn={978-3-540-12732-1},
  doi={10.1007/3-540-12732-1}
}

@article{belov2024partiallyPRE,
  title = {{Partially unitary learning}},
  author = {Belov, Mikhail Gennadievich and Malyshkin, Vladislav Gennadievich},
  journal = {Phys. Rev. E},
  volume = {110},
  issue = {5},
  pages = {055306},
  numpages = {18},
  year = {2024},
  month = {Nov},
  publisher = {American Physical Society},
  doi = {10.1103/PhysRevE.110.055306},
  url = {https://link.aps.org/doi/10.1103/PhysRevE.110.055306}
}

@article{choi1975completely,
  title={{Completely positive linear maps on complex matrices}},
  author={Choi, Man-Duen},
  journal={Linear Algebra and its Applications},
  volume={10},
  number={3},
  pages={285--290},
  year={1975},
  publisher={Elsevier},
  doi={10.1016/0024-3795(75)90075-0}
}

@article{belavkin1986radon,
  title={{A Radon-Nikodym theorem for completely positive maps}},
  author={Belavkin, Viacheslav P and Staszewski, Przemyslaw},
  journal={Reports on Mathematical Physics},
  volume={24},
  number={1},
  pages={49--55},
  year={1986},
  publisher={Elsevier},
  doi={10.1016/0034-4877(86)90039-X}
}

@article{belov2024quantumPRE,
  title = {{Quantum channel learning}},
  author = {Belov, Mikhail Gennadievich and Dubov, Victor Victorovich and Filimonov, Alexey Vladimirovich and Malyshkin, Vladislav Gennadievich},
  journal = {Phys. Rev. E},
  volume = {111},
  issue = {1},
  pages = {015302},
  numpages = {20},
  year = {2025},
  month = {Jan},
  publisher = {American Physical Society},
  doi = {10.1103/PhysRevE.111.015302},
  url = {https://link.aps.org/doi/10.1103/PhysRevE.111.015302}
}

@article{birman2004ija,
  title={{Ija Pavlovna Ipatova}},
  author={Birman, Joseph L and Maradudin, Alexei A and Pick, Robert and Rebane, Karl K},
   journal = {Physics Today},
    volume = {57},
    number = {8},
    pages = {69-70},
    year = {2004},
    month = {08},
  publisher={American Institute of Physics},
  doi={10.1063/1.1801876}
}

@article{jamiolkowski1972linear,
  title={{Linear transformations which preserve trace and positive semidefiniteness of operators}},
  author={Jamio{\l}kowski, Andrzej},
  journal={Reports on Mathematical Physics},
  volume={3},
  number={4},
  pages={275--278},
  year={1972},
  publisher={Elsevier},
  doi={10.1016/0034-4877(72)90011-0}
}

@article{torlai2023quantum,
  title={{Quantum process tomography with unsupervised learning and tensor networks}},
  author={Torlai, Giacomo and Wood, Christopher J and Acharya, Atithi and Carleo, Giuseppe and Carrasquilla, Juan and Aolita, Leandro},
  journal={Nature Communications},
  volume={14},
  number={1},
  pages={2858},
  year={2023},
  publisher={Nature Publishing Group UK London},
  doi={10.1038/s41467-023-38332-9}
}

@article{belov2025superstatePRE,
  title={{Superstate Quantum Mechanics}},
  author={Belov, Mikhail Gennadievich and Dubov, Victor Victorovich and Ivanov, Vadim Konstantinovich and Maslov, Alexander Yurievich and Proshina, Olga Vladimirovna and Malyshkin, Vladislav Gennadievich},
  journal={arXiv preprint arXiv:2502.00037},
  year={2025},
  doi={10.48550/arXiv.2502.00037}
}

@article{belov2025tradePRE,
  title={{Trade Execution Flow as the Underlying Source of Market Dynamics}},
  author={Belov, Mikhail Gennadievich and Dubov, Victor Victorovich and Ivanov, Vadim Konstantinovich and Maslov, Alexander Yurievich and Proshina, Olga Vladimirovna and Malyshkin, Vladislav Gennadievich},
  journal={arXiv preprint arXiv:2511.01471},
  year={2025},
  doi={10.48550/arXiv.2511.01471}
}

@MISC{EP12456supplemental,
author={Malyshkin, Vladislav Gennadievich},
title={{The supplimental materials for Phys. Rev. E \textbf{114}, 035302 (2026) \doi{10.1103/kdl5-smrh}}},
note={Includes the software implementing our algorithms, build scripts, unit tests,
and integration code for commercial SDP software needed to reproduce the results,
available at \url{http://link.aps.org/supplemental/10.1103/kdl5-smrh}},
year="2026",
url={http://link.aps.org/supplemental/10.1103/kdl5-smrh}
}

@MISC{iipColloq2025,
author={Malyshkin, Vladislav Gennadievich},
title={{Quantum Channel as Classical Computational Model}},
note={{Scientific Colloquium in Honor of the 96th Birthday of I. P. Ipatova.
Held on December 19, 2025.}
\url{http://www.ioffe.ru/LNEPS/abstracts/ipatova_96.html}},
year = {2025},
month = {Dec},
day = {19},
url={http://www.ioffe.ru/LNEPS/abstracts/ipatova_96.html}
}

@book{vapnik2013nature,
  title={{The nature of statistical learning theory}},
  author={Vapnik, Vladimir},
  year={2013},
  publisher={Springer Science \& Business Media},
  doi={10.1007/978-1-4757-3264-1}
}

@article{bengio2013representation,
  title={{Representation learning: A review and new perspectives}},
  author={Bengio, Yoshua and Courville, Aaron and Vincent, Pascal},
  journal={IEEE transactions on pattern analysis and machine intelligence},
  volume={35},
  number={8},
  pages={1798--1828},
  year={2013},
  publisher={IEEE},
  doi={10.1109/TPAMI.2013.50}
}

@article{witten2002data,
  title={{Data mining: practical machine learning tools and techniques with Java implementations}},
  author={Witten, Ian H and Frank, Eibe},
  journal={Acm Sigmod Record},
  volume={31},
  number={1},
  pages={76--77},
  year={2002},
  publisher={ACM New York, NY, USA},
  doi={10.1016/C2009-0-19715-5}
}

@article{vapnik1974method,
  title={{The method of ordered risk minimization, I}},
  author={Vapnik, VN and Chervonenkis, A Ya},
  journal={Avtomatika i Telemekhanika},
  volume={8},
  pages={21--30},
  year={1974},
  url={http://www.mathnet.ru/php/archive.phtml?wshow=paper&jrnid=at&paperid=8452&option_lang=eng}
}

@article{hajek1977generation,
  title={{On generation of inductive hypotheses}},
  author={H{\'a}jek, Petr and Havr{\'a}nek, Tom{\'a}{\v{s}}},
  journal={International Journal of Man-Machine Studies},
  volume={9},
  number={4},
  pages={415--438},
  year={1977},
  publisher={Elsevier},
  doi={10.1016/S0020-7373(77)80011-4}
}

@article{zadeh1965fuzzy,
  title={{Fuzzy sets}},
  author={Zadeh, Lotfi A},
  journal={Information and Control},
  volume={8},
  number={3},
  pages={338--353},
  year={1965},
  publisher={Elsevier},
  doi={10.1016/S0019-9958(65)90241-X}
}

@article{hajek1995fuzzy,
  title={{Fuzzy logic and arithmetical hierarchy}},
  author={H{\'a}jek, Petr},
  journal={Fuzzy Sets and Systems},
  volume={73},
  number={3},
  pages={359--363},
  year={1995},
  publisher={Elsevier},
  doi={10.1016/0165-0114(94)00299-M}
}

@inproceedings{arjovsky2016unitary,
  title={{Unitary evolution recurrent neural networks}},
  author={Arjovsky, Martin and Shah, Amar and Bengio, Yoshua},
  booktitle={International conference on machine learning, NY, USA, 2016},
  pages={1120--1128},
  year={2016},
  organization={Proceedings of Machine Learning Research (proceedings.mlr.press) },
  doi={10.48550/arXiv.1511.06464}
}

@article{bisio2010optimal,
  title={{Optimal quantum learning of a unitary transformation}},
  author={Bisio, Alessandro and Chiribella, Giulio and D’Ariano, Giacomo Mauro and Facchini, Stefano and Perinotti, Paolo},
  journal={Phys. Rev. A},
  volume={81},
  number={3},
  pages={032324},
  year={2010},
  publisher={APS},
  doi={10.1103/PhysRevA.81.032324}
}

@inproceedings{hyland2017learning,
  title={{Learning unitary operators with help from u(n)}},
  author={Hyland, Stephanie and R{\"a}tsch, Gunnar},
  booktitle={Proceedings of the AAAI Conference on Artificial Intelligence},
  volume={31},
  number={1},
  year={2017},
  organization={Association for the Advancement of Artificial Intelligence (aaai.org)},
  doi={10.1609/aaai.v31i1.10928}
}

@book{nielsen2010quantum,
  title={Quantum computation and quantum information},
  author={Nielsen, Michael A and Chuang, Isaac L},
  year={2010},
  publisher={Cambridge University Press},
  doi={10.1017/CBO9780511976667}
}

@article{wilde2011classical,
  title={{From classical to quantum Shannon theory}},
  author={Wilde, Mark M},
  journal={arXiv preprint arXiv:1106.1445},
  year={2011},
  doi={10.48550/arXiv.1106.1445}
}

@article{huang2021learning,
  title={{Learning Unitary Transformation by Quantum Machine Learning Model.}},
  author={Huang, Yi-Ming and Li, Xiao-Yu and Zhu, Yi-Xuan and Lei, Hang and Zhu, Qing-Sheng and Yang, Shan},
  journal={Computers, Materials \& Continua},
  volume={68},
  number={1},
  year={2021},
  doi={10.32604/cmc.2021.016663}
}

@article{yu2023optimal,
  title={{Optimal quantum dataset for learning a unitary transformation}},
  author={Yu, Zhan and Zhao, Xuanqiang and Zhao, Benchi and Wang, Xin},
  journal={Phys. Rev. Applied},
  volume={19},
  number={3},
  pages={034017},
  year={2023},
  publisher={APS},
  doi={10.1103/PhysRevApplied.19.034017}
}

@article{wen2013feasible,
  title={{A feasible method for optimization with orthogonality constraints}},
  author={Wen, Zaiwen and Yin, Wotao},
  journal={Mathematical Programming},
  volume={142},
  number={1},
  pages={397--434},
  year={2013},
  publisher={Springer},
  doi={10.1007/s10107-012-0584-1}
}

@article{budini2024quantum,
  title={{Quantum distinguishability measures: Projectors versus state maximization}},
  author={Budini, Adri{\'a}n A and Filho, Ruynet L {\relax de} Matos and Santos, Marcelo F},
  journal={Phys. Rev. A},
  volume={110},
  number={2},
  pages={022228},
  year={2024},
  publisher={APS},
  doi={10.1103/PhysRevA.110.022228}
}

@article{holzapfel2015scalable,
  title={{Scalable reconstruction of unitary processes and Hamiltonians}},
  author={Holz{\"a}pfel, M and Baumgratz, T and Cramer, M and Plenio, Martin B},
  journal={Phys. Rev. A},
  volume={91},
  number={4},
  pages={042129},
  year={2015},
  publisher={APS},
  doi={10.1103/PhysRevA.91.042129}
}

@article{taranto2025higher,
  title={{Higher-Order Quantum Operations}},
  author={Taranto, Philip and Milz, Simon and Murao, Mio and Quintino, Marco T{\'u}lio and Modi, Kavan},
  journal={arXiv preprint arXiv:2503.09693},
  year={2025},
  doi={10.48550/arXiv.2503.09693}
}

@article{baldwin2014quantum,
  title={{Quantum process tomography of unitary and near-unitary maps}},
  author={Baldwin, Charles H and Kalev, Amir and Deutsch, Ivan H},
  journal={Phys. Rev. A},
  volume={90},
  number={1},
  pages={012110},
  year={2014},
  publisher={APS},
  doi={10.1103/PhysRevA.90.012110}
}

@inproceedings{wilde2018recoverability,
  title={{Recoverability for Holevo's just-as-good fidelity}},
  author={Wilde, Mark M},
  booktitle={2018 IEEE International Symposium on Information Theory (ISIT)},
  pages={2331--2335},
  year={2018},
  organization={IEEE},
  doi={10.1109/ISIT.2018.8437346}
}

@book{anjos2011handbook,
  title={{Handbook on semidefinite, conic and polynomial optimization}},
  author={Anjos, Miguel F and Lasserre, Jean B},
  volume={166},
  year={2011},
  publisher={Springer Science \& Business Media},
  doi={10.1007/978-1-4614-0769-0}
}

@article{johnston2011quantum,
  title={{Quantum gate fidelity in terms of Choi matrices}},
  author={Johnston, Nathaniel and Kribs, David W},
  journal={Journal of Physics A: Mathematical and Theoretical},
  volume={44},
  number={49},
  pages={495303},
  year={2011},
  publisher={IOP Publishing},
  doi={10.1088/1751-8113/44/49/495303}
}

@article{vandenberghe1996semidefinite,
  title={{Semidefinite programming}},
  author={Vandenberghe, Lieven and Boyd, Stephen},
  journal={SIAM Review},
  volume={38},
  number={1},
  pages={49--95},
  year={1996},
  publisher={SIAM},
  doi={10.1137/1038003}
}

@book{wolkowicz2012handbook,
  title={{Handbook of semidefinite programming: theory, algorithms, and applications}},
  author={Wolkowicz, Henry and Saigal, Romesh and Vandenberghe, Lieven},
  volume={27},
  year={2012},
  publisher={Springer Science \& Business Media},
  doi={10.1007/978-1-4615-4381-7}
}

@book{boyd2004convex,
  title={Convex optimization},
  author={Boyd, Stephen and Vandenberghe, Lieven},
  year={2004},
  publisher={Cambridge University Press},
  doi={10.1017/CBO9780511804441}
}

@article{yamashita2010high,
  title={{A high-performance software package for semidefinite programs: SDPA 7}},
  author={Yamashita, Makoto and Fujisawa, Katsuki and Nakata, Kazuhide and Nakata, Maho and Fukuda, Mituhiro and Kobayashi, Kazuhiro and Goto, Kazushige},
  journal={Tokyo, Japan},
  year={2010},
  url={https://optimization-online.org/wp-content/uploads/2010/01/2531.pdf}
}

@article{diamond2016cvxpy,
  author = {Steven Diamond and Stephen Boyd},
  title = {{CVXPY}: {A} {P}ython-embedded modeling language for convex optimization},
  journal = {Journal of Machine Learning Research},
  year = {2016},
  volume = {17},
  number = {83},
  pages = {1--5},
  url={https://www.jmlr.org/papers/volume17/15-408/15-408.pdf}
}

@misc{andersen2015cvxopt,
  author = {Andersen, M. and Dahl, J. and Vandenberghe, L.},
  title = {{CVXOPT}: {P}ython software for convex optimization},
  howpublished = {\url{http://cvxopt.org/}},
  year = {2015},
  note = {Version 1.7},
  url={http://cvxopt.org/}
}

@book{skrzypczyk2023semidefinite,
  title={{Semidefinite programming in quantum information science}},
  author={Skrzypczyk, Paul and Cavalcanti, Daniel},
  year={2023},
  publisher={IOP Publishing},
  doi={10.1088/978-0-7503-3343-6}
}

@article{tavakoli2024semidefinite,
  title={{Semidefinite programming relaxations for quantum correlations}},
  author={Tavakoli, Armin and Pozas-Kerstjens, Alejandro and Brown, Peter and Ara{\'u}jo, Mateus},
  journal={Reviews of Modern Physics},
  volume={96},
  number={4},
  pages={045006},
  year={2024},
  publisher={APS},
  doi = {10.1103/RevModPhys.96.045006}
}

@book{schuld2021machine,
  title={{Machine learning with quantum computers}},
  author={Schuld, Maria and Petruccione, Francesco},
  volume={676},
  year={2021},
  publisher={Springer},
  doi={10.1007/978-3-030-83098-4}
}

@article{biamonte2017quantum,
  title={{Quantum machine learning}},
  author={Biamonte, Jacob and Wittek, Peter and Pancotti, Nicola and Rebentrost, Patrick and Wiebe, Nathan and Lloyd, Seth},
  journal={Nature},
  volume={549},
  number={7671},
  pages={195--202},
  year={2017},
  publisher={Nature Publishing Group UK London},
  doi={10.1038/nature23474}
}

\end{document}